\documentclass{article} 
\usepackage{iclr2027_conference,times}
\usepackage{hyperref}
\usepackage{url}
\usepackage{booktabs}
\usepackage[table]{xcolor}
\usepackage{graphicx}
\usepackage{wrapfig}
\usepackage{subcaption}

\usepackage{amsmath,amsfonts,bm}

\def\eqref#1{equation~\ref{#1}}

\def\1{\bm{1}}

\DeclareMathAlphabet{\mathsfit}{\encodingdefault}{\sfdefault}{m}{sl}
\SetMathAlphabet{\mathsfit}{bold}{\encodingdefault}{\sfdefault}{bx}{n}

\title{AdaptiveEmbed: Sample-Adaptive Multi-Vector Representation for Multimodal Retrieval}

\author{
Xinze Liu$^{1,2\,*}$\quad
Lei Yang$^{1,2\,*}$\quad
Dayan Wu$^{1\,\dagger}$\quad
Hengjie Zhu$^{1,2}$\quad
Zihao Zhang$^{1,2}$\quad
Hanqi Wu$^{1,2}$ \\
Tianzhu Hu$^{1,2}$\quad
Peng Fu$^{1}$\quad
Zheng Lin$^{1}$\quad
Weiping Wang$^{1}$ \\
\vspace{2pt} \\
$^{1}$Institute of Information Engineering, Chinese Academy of Sciences \\
$^{2}$School of Cyber Security, University of Chinese Academy of Sciences \\
\texttt{\{liuxinze,wudayan\}@iie.ac.cn}
}

\iclrfinalcopy
\begin{document}
\maketitle
\renewcommand{\thefootnote}{\fnsymbol{footnote}}
\footnotetext[1]{Equal contribution.}
\footnotetext[2]{Corresponding author.}
\renewcommand{\thefootnote}{\arabic{footnote}}
\begin{abstract}
Multi-vector representations have emerged as an effective paradigm for multimodal retrieval, representing each sample with multiple complementary embeddings to capture fine-grained cross-modal information. However, existing approaches typically employ a fixed representation capacity, assigning the same number of vectors to all samples regardless of their individual retrieval demands. Such a fixed-capacity formulation overlooks the fact that different samples may require different amounts of representation capacity for effective retrieval. In this work, we introduce \emph{Sample-Adaptive Multi-Vector Representation} (SAMVR), a new problem setting for multimodal retrieval that studies how multi-vector representation capacity can be allocated at the sample level. Under SAMVR, each sample is represented by a \emph{content-adaptive embedding set} (CAES), whose capacity is determined according to the sample-specific retrieval utility of additional representation vectors. To instantiate SAMVR, we propose \emph{AdaptiveEmbed}, a unified framework for learning sample-adaptive multi-vector representations. AdaptiveEmbed learns structured multi-vector representations through \emph{Multi-Group Contrastive Learning} (MGCL) with the symmetric \emph{set-to-set similarity} (SetSim), and further employs \emph{Utility Policy Optimization} (UPO) to determine sample-specific representation capacity via \emph{Marginal Utility Allocation} (MUA). Experiments across multimodal retrieval benchmarks involving image, text, video, and audio show that sample-adaptive capacity allocation achieves overall better retrieval performance than fixed-capacity multi-vector representations, validating the effectiveness of SAMVR for multimodal retrieval. These results establish SAMVR as a viable formulation for adaptive capacity allocation in multi-vector multimodal retrieval.
\end{abstract}

\section{Introduction}

Multimodal retrieval has increasingly moved from compact single-vector embeddings toward more expressive multi-vector representations. Single-vector models provide an efficient comparison interface by compressing each sample into a single embedding \citep{radford2021clip,zhai2023siglip}, whereas multi-vector representations expose multiple embeddings to the matching function and thereby support richer fine-grained interactions across modalities \citep{khattab2020colbert,santhanam2022colbertv2,faysse2025colpali}. Existing studies have primarily focused on how multiple embeddings should be constructed, organized, and compared. In contrast, considerably less attention has been paid to another dimension of the representation: \emph{whether different samples should use the same representation capacity}.

Most existing multi-vector retrieval methods adopt a largely sample-independent representation budget. Learned multi-vector encoders commonly produce a fixed number of representation vectors, while token- or region-level approaches often inherit their representation size from the underlying input structure. More recent flexible representations support multiple operating points at inference time, allowing the retrieval system to trade representation capacity for retrieval cost \citep{xiao2026metaembed}. However, such operating points are typically selected at the system level rather than independently for each sample. Consequently, the amount of multi-vector capacity exposed to retrieval remains largely uniform across samples under a given operating configuration.

\begin{figure}[t]
\centering
\includegraphics[width=\linewidth]{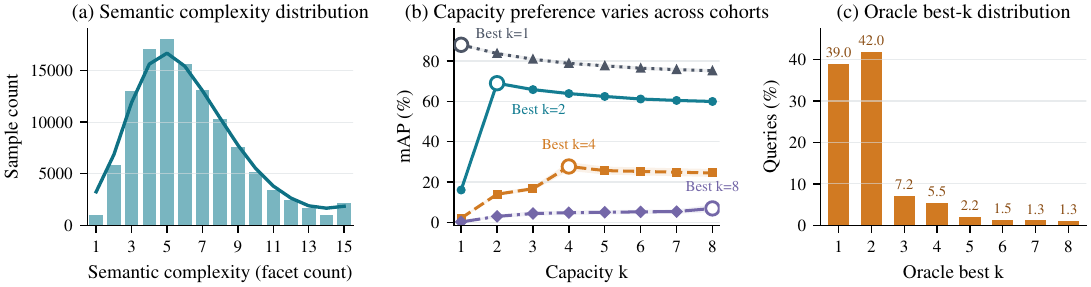}
\caption{\textbf{Motivation for sample-adaptive multi-vector representation in multimodal retrieval.}
(a) Multimodal samples exhibit diverse content characteristics.
(b) Samples grouped by oracle-optimal $k$ show different capacity--performance trends, motivating sample-adaptive allocation.
(c) Oracle-preferred capacities are distributed across the available range, indicating substantial sample-level variation in representation demand.
Results are visualized on COCO.}
\label{fig:motivation}
\end{figure}

We investigate whether this uniform allocation is necessary. Figure~\ref{fig:motivation} presents three complementary observations. First, multimodal samples exhibit substantial content heterogeneity (Fig.~\ref{fig:motivation}(a)), suggesting that a single representation budget may be restrictive. However, we find that such surface-level content complexity does not by itself determine how much capacity a sample actually needs; we therefore characterize capacity through its \emph{retrieval utility} rather than any explicit complexity measure. Grouping samples by their oracle-optimal $k$ (Fig.~\ref{fig:motivation}(b)) reveals markedly different capacity--performance trends across samples, and at the dataset level the oracle-preferred capacities are spread across the entire available range (Fig.~\ref{fig:motivation}(c)). Together, these observations indicate that the capacity most useful for retrieval varies substantially from one sample to another.

This motivates a different formulation of multi-vector multimodal retrieval: representation capacity itself can be treated as a sample-dependent decision variable. The objective is not simply to expose more vectors, nor to tie capacity to a predefined notion of content complexity, but to allocate representation capacity according to its retrieval utility for each sample. Under this view, multi-vector retrieval moves beyond a globally configured representation budget toward \emph{sample-adaptive capacity allocation}.

We formalize this problem as \textbf{Sample-Adaptive Multi-Vector Representation (SAMVR)}, a new setting for multimodal retrieval in which representation capacity is allocated at the sample level. Given an individual sample, SAMVR asks how much multi-vector capacity should be assigned to it and how the corresponding representation should be constructed according to the utility of additional embeddings. We refer to the resulting representation as a \emph{content-adaptive embedding set} (CAES), whose composition and capacity are determined individually for each sample.

To instantiate SAMVR, we propose \emph{AdaptiveEmbed}, a unified framework for learning sample-adaptive multi-vector representations in multimodal retrieval. AdaptiveEmbed separates the learning of candidate representations from the decision of how much representation capacity to allocate to each sample. It first constructs structured candidate embeddings through \emph{Multi-Group Contrastive Learning} (MGCL) with the symmetric \emph{set-to-set similarity} (SetSim). It then introduces \emph{Utility Policy Optimization} (UPO) to learn sample-specific capacity decisions, together with \emph{Marginal Utility Allocation} (MUA) for allocating additional representations according to their benefit. Together, these components provide a practical realization of SAMVR and enable adaptive multi-vector retrieval within a unified framework.

Our contributions are threefold: \textbf{(i)} we introduce Sample-Adaptive Multi-Vector Representation (SAMVR), a new problem setting for multimodal retrieval that treats multi-vector capacity as a per-sample decision variable driven by retrieval utility; \textbf{(ii)} we propose AdaptiveEmbed, a unified framework instantiating SAMVR that couples multi-group representation learning, a symmetric set-to-set similarity (SetSim) allowing a single embedding set to serve both retrieval directions, and a utility-driven policy that allocates capacity per sample; and \textbf{(iii)} across image, text, video, and audio benchmarks, sample-adaptive allocation achieves higher average mAP than fixed-capacity multi-vector representations while using only around two embeddings per sample versus \texttt{3--40} for fixed-capacity baselines.


\section{Related Work}
\paragraph{Multi-Vector Retrieval.} Dense retrieval compresses each input into a single embedding, enabling efficient search but limiting the fine-grained evidence retained in the representation \citep{karpukhin2020dpr,radford2021clip}. Multi-vector retrieval alleviates this by representing each sample with multiple embeddings and modeling richer interactions among them. ColBERT established token-level late interaction through MaxSim \citep{khattab2020colbert}, followed by work improving interaction quality and efficiency and extending multi-vector representations to multimodal retrieval, where ColPali exposes patch-level visual embeddings and MetaEmbed introduces nested embedding sets with flexible test-time vector budgets \citep{santhanam2022colbertv2,faysse2025colpali,xiao2026metaembed}. A parallel line reduces the storage and computation overhead of multi-vector retrieval through nested representations, fixed-dimensional encodings, and efficient late-interaction engines \citep{kusupati2022matryoshka,dhulipala2024muvera}. These methods make multi-vector retrieval increasingly expressive and scalable, but operate under predefined representation structures or globally selected capacity budgets, leaving the per-sample allocation of capacity unexplored.

\paragraph{Retrieval Feedback for Adaptive Decisions.} Retrieval results themselves can provide query-dependent feedback about the retrieval environment. Pseudo-relevance feedback refines query representations from initially retrieved candidates, including extensions to multi-vector dense retrieval \citep{wang2021prf}, while selective feedback methods learn when such refinement should be applied \citep{datta2024selective}. More recently, Col-Bandit shows that the MaxSim interaction matrix itself carries informative query-dependent signals for adaptive late-interaction computation \citep{pony2026colbandit}. These studies motivate retrieval feedback as an observable signal beyond the query representation, which we exploit to guide per-sample capacity decisions.

SAMVR studies a complementary problem: rather than constructing more vectors, compressing a predefined representation, or choosing a global operating budget, we treat the \emph{composition and capacity} of the embedding set as per-sample variables. Each sample progressively selects representations according to their retrieval utility, with MaxSim responses from a retrieval bank serving as observable feedback. In this sense, SAMVR shifts multi-vector retrieval from a globally configured budget toward \emph{sample-adaptive capacity allocation}.

\section{Methodology}
We first formulate multimodal retrieval with sample-adaptive multi-vector representations and introduce SetSim as the set-to-set similarity. We then present AdaptiveEmbed, covering its architecture, multi-group representation learning, and utility-driven capacity allocation.
\subsection{Preliminaries}
\paragraph{Problem Definition.} Let $\mathcal{X}^{\mathcal{A}}$ and $\mathcal{X}^{\mathcal{B}}$ denote the sample spaces of two modalities. Given a query $x\in\mathcal{X}^{\mathcal{A}}$ and a cross-modal gallery $\mathcal{G}^{\mathcal{B}}=\{y_j\}_{j=1}^{M}\subset\mathcal{X}^{\mathcal{B}}$, multimodal retrieval ranks the gallery samples according to their similarity to $x$; retrieval in the reverse direction is defined analogously. SAMVR maps each sample $x\in\mathcal{X}^{\mathcal{A}}\cup\mathcal{X}^{\mathcal{B}}$ to a sample-specific multi-vector representation:
\begin{equation}
f_{\theta}:x\mapsto
Z_x=\{z_{x,i}\}_{i=1}^{n_x},
\qquad
z_{x,i}\in\mathbb{R}^{d},
\quad
n_x\in\mathcal{K},
\label{eq:samvr_mapping}
\end{equation}
where $Z_x$ is the content-adaptive embedding set (CAES), $n_x$ is its sample-specific capacity, and $\mathcal{K}$ denotes the set of available capacities. Throughout, we use $N$ for the size of the candidate embedding pool and $k$ for an activated capacity, $k\in\mathcal{K}$. We use $o\in\{\mathcal{A}\!\rightarrow\!\mathcal{B},\,\mathcal{B}\!\rightarrow\!\mathcal{A}\}$ to index the two retrieval directions.
\begin{figure}[t]
\centering
\includegraphics[width=\linewidth]{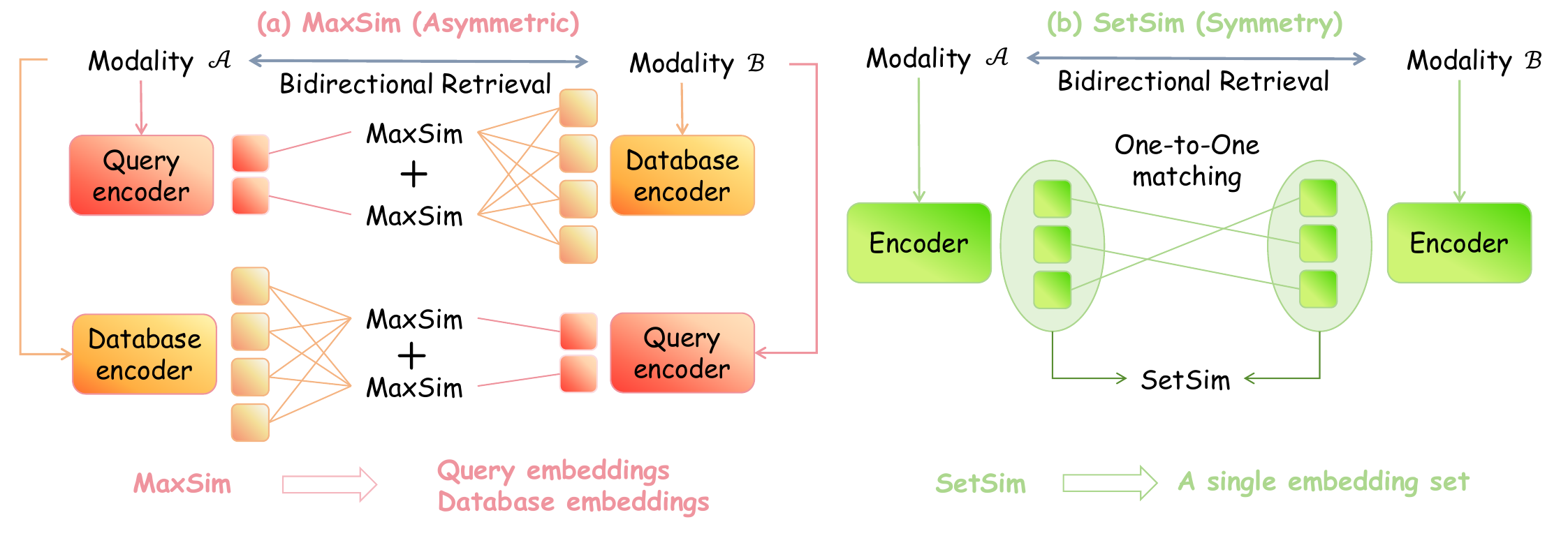}
\caption{
Comparison between MaxSim and SetSim for bidirectional multimodal retrieval.
(a) MaxSim defines a directed comparison, leading the two retrieval directions to use role-specific query/database configurations.
(b) SetSim performs symmetric one-to-one matching between embedding sets, allowing each modality to use a single embedding set for both retrieval directions.
}
\label{fig:setsim}
\end{figure}

\paragraph{Set-to-Set Similarity.}
\label{sec:setsim}
Figure~\ref{fig:setsim} contrasts MaxSim and SetSim in bidirectional multimodal retrieval. MaxSim independently matches each query-side vector to its most similar candidate-side vector, so exchanging the two sets changes the matching direction. SetSim instead applies the same set-level comparison in both directions, allowing the same embedding set to represent a sample regardless of the retrieval direction.

Given two L2-normalized embedding sets of equal cardinality, $Q=\{q_i\}_{i=1}^{k}$ and $D=\{d_j\}_{j=1}^{k}$, their token-level similarity matrix is defined by $C_{ij}=q_i^{\top}d_j$. SetSim finds the bijective correspondence with the maximum average similarity:
\begin{equation}
\operatorname{SetSim}(Q,D)=
\frac{1}{k}
\max_{\pi\in\Pi_k}
\sum_{i=1}^{k} C_{i,\pi(i)},
\label{eq:setsim}
\end{equation}
where $\Pi_k$ is the set of permutations over $k$ elements, and $\pi$ assigns each vector in $Q$ to a unique vector in $D$, i.e., a maximum-weight one-to-one assignment~\citep{setsim}. Since the assignment is a bijection between two equal-cardinality sets, the resulting score is symmetric in $Q$ and $D$. With at most $N$ tokens per set, we compute the optimal matching exactly rather than approximately; the same exact computation is used in both training and inference.

\textbf{Retrieval-time activation.}
At retrieval time, the query's CAES $Z_x$ determines the activation granularity: it specifies which token positions are active, and each gallery sample activates the \emph{same} positions of its embedding set for comparison. Because the token positions are aligned across samples through the group-prefix structure learned in Stage~I (Sec.~\ref{sec:mgcl}), this yields two equal-cardinality sets whose coarse- and fine-grained tokens are matched in kind, so SetSim in Eq.~\ref{eq:setsim} is always applied between sets of equal size.
\begin{figure}[t]
\centering
\includegraphics[width=\linewidth]{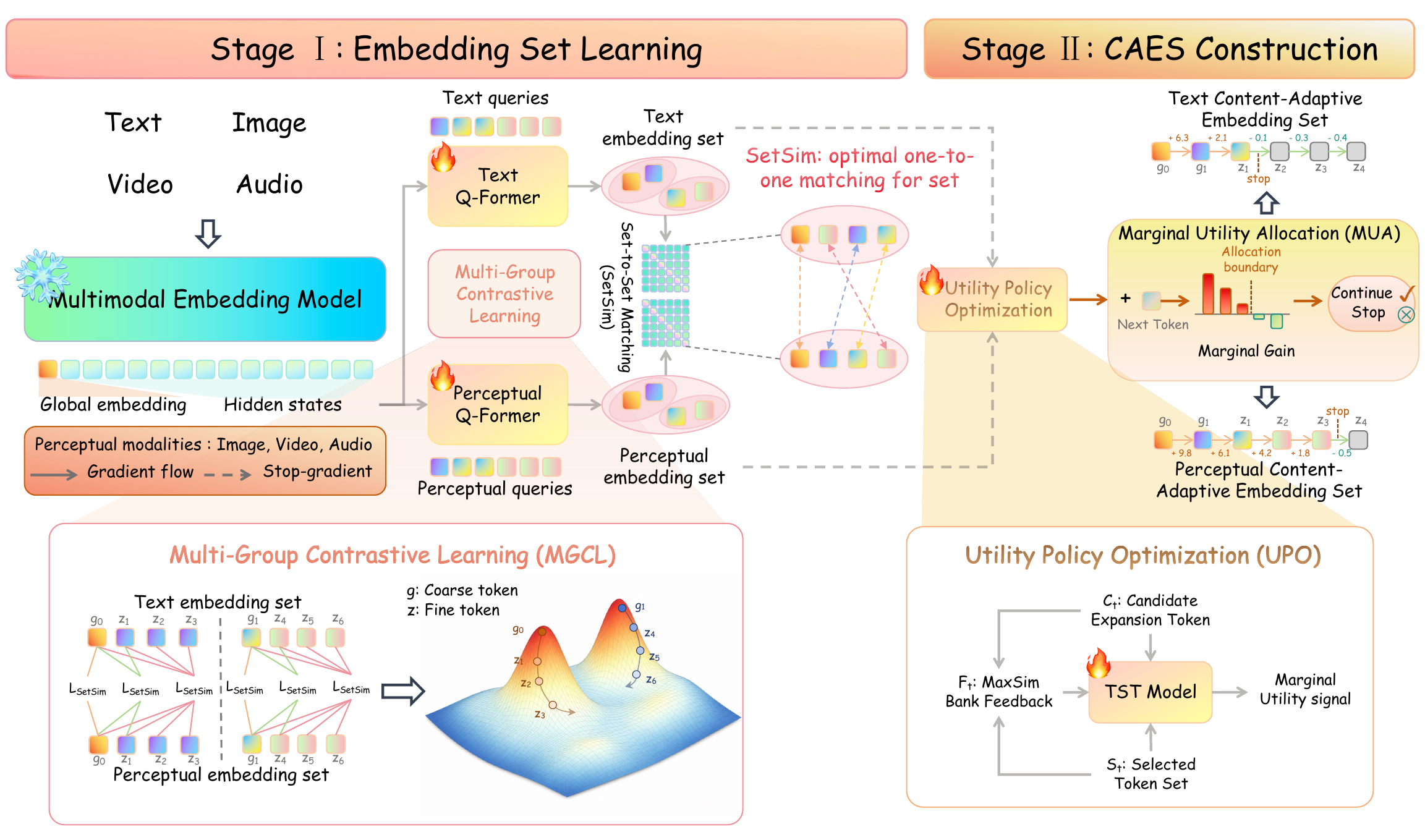}
\caption{Overview of AdaptiveEmbed. The framework first learns a structured multi-vector embedding set through Multi-Group Contrastive Learning (MGCL), and then constructs a sample-specific content-adaptive embedding set (CAES) using a Token Selection Transformer (TST) trained with Utility Policy Optimization (UPO) and a Marginal Utility Allocation (MUA) strategy.}
\label{fig:framework}
\end{figure}
\subsection{Architecture}
\label{subsec:arc}
Figure~\ref{fig:framework} presents the two-stage architecture of AdaptiveEmbed. Stage~I learns a structured, fixed-size pool of candidate embeddings at multiple capacity levels. Stage~II evaluates the retrieval utility of expanding the current representation and selects a sample-specific subset as the final content-adaptive embedding set (CAES). For retrieval, the CAES of a query is compared with each gallery CAES using SetSim, and the gallery is ranked by the resulting similarity scores.

\paragraph{Stage I: Embedding Set Learning.}
Let $x\in\mathcal{X}^{(m)}$ denote a sample from modality $m$. Its corresponding modality encoder produces a global feature $g_x\in\mathbb{R}^{d_e}$ and hidden states $H_x\in\mathbb{R}^{L_x\times d_e}$, where $L_x$ is the input sequence length and $d_e$ is the encoder dimension. 
A Q-Former~\citep{li2023blip2} processes $H_x$ using $N$ learnable queries and outputs $Q_x=[q_{x,1},\ldots,q_{x,N}]\in\mathbb{R}^{N\times d_q}$. Here $d_e$, $d_q$, and $d$ denote the encoder, Q-Former, and shared retrieval dimensions, respectively, and $P^{(m)}$ maps from $d_q$ to $d$. Each output is projected into the shared retrieval space by a modality-specific projection $P^{(m)}\in\mathbb{R}^{d\times d_q}$ and L2-normalized as $e_{x,i}=P^{(m)}q_{x,i}/\|P^{(m)}q_{x,i}\|_2$. The resulting candidate embedding set is denoted by $E_x=\{e_{x,i}\}_{i=1}^{N}$, with $e_{x,i}\in\mathbb{R}^{d}$.

\paragraph{Multi-Group Contrastive Learning (MGCL).} 
\label{sec:mgcl}
MGCL organizes $E_x$ into $G$ ordered groups. 
The $g$-th group is written as $E_{x,g}=\{e_{x,g,r}\}_{r=1}^{K_g}$, where $K_g$ is its size and $\sum_{g=1}^{G}K_g=N$. Each group follows an ordered prefix structure: for a supported capacity $k\in\mathcal{K}_g\subseteq\{1,\ldots,K_g\}$, its $k$-vector representation is $E_{x,g}^{(k)}=\{e_{x,g,r}\}_{r=1}^{k}$. Following Matryoshka Representation Learning and group-based multi-vector learning~\citep{kusupati2022matryoshka,xiao2026metaembed}, MGCL jointly optimizes all supported prefixes so that shorter prefixes provide compact representations and longer prefixes introduce additional retrieval information.

Given a paired minibatch
$\mathcal{B}=\{(x_n^{\mathcal{A}},x_n^{\mathcal{B}})\}_{n=1}^{N_b}$,
we define
$s_{n,m}^{g,k}
=\operatorname{SetSim}(E_{x_n^{\mathcal{A}},g}^{(k)},
E_{x_m^{\mathcal{B}},g}^{(k)})$
as the similarity between the $n$-th sample of modality $\mathcal{A}$ and the $m$-th sample of modality $\mathcal{B}$. Let
$N_{\mathrm{p}}=\sum_{g=1}^{G}|\mathcal{K}_g|$
denote the total number of optimized group-prefix configurations. MGCL applies bidirectional InfoNCE~\citep{oord2018cpc} to every configuration:
\begin{equation}
\mathcal{L}_{\mathrm{MGCL}}
=
-\frac{1}{2N_bN_{\mathrm{p}}}
\sum_{g=1}^{G}
\sum_{k\in\mathcal{K}_g}
\sum_{n=1}^{N_b}
\left[
\log
\frac{\exp(s_{n,n}^{g,k}/\tau)}
{\sum_{m=1}^{N_b}\exp(s_{n,m}^{g,k}/\tau)}
+
\log
\frac{\exp(s_{n,n}^{g,k}/\tau)}
{\sum_{m=1}^{N_b}\exp(s_{m,n}^{g,k}/\tau)}
\right],
\label{eq:mgcl}
\end{equation}
where $\tau>0$ is the temperature. The two terms optimize retrieval from modality $\mathcal{A}$ to modality $\mathcal{B}$ and in the reverse direction, respectively. The resulting ordered groups constitute the candidate representation space used by Stage~II.
\paragraph{Stage II: CAES Construction.}
Stage~II determines which candidate embeddings should be allocated to each sample. At allocation step $t$, let $\mathcal{S}_t\subseteq E_x$ denote the currently selected embeddings and let $\mathcal{C}_t\subseteq E_x\setminus\mathcal{S}_t$ contain the admissible next embeddings, i.e., those that extend one of the ordered groups by its next prefix element while keeping every group a valid prefix (Appendix~\ref{app:adaptive_allocation}). 
We use a fixed retrieval bank $\mathcal{R}$ to provide the policy with an observation of how the current representation behaves during retrieval. For a query token $q_i$ and a bank item $D_n$, we compute its MaxSim response $A_{i,n}=\max_{d\in D_n} q_i^\top d$; ranking bank items by the response of the currently selected tokens $\mathcal{S}_t$ and keeping the top ones yields a feedback matrix $\mathcal{F}_t$ that records how both selected and candidate tokens respond to the most relevant bank items, following the observation that MaxSim $Top~L$ responses carry informative query-dependent feedback~\citep{pony2026colbandit}. This feedback adds no retrieved content to the representation, and exposes neither ground-truth identities nor retrieval ranks or utilities to the policy; the marginal utilities in Eq.~\ref{eq:upo} are detached and used only as training targets. The allocation state is therefore $s_t=(\mathcal{S}_t,\mathcal{C}_t,\mathcal{F}_t)$, and its exact construction and encoding are detailed in Appendix~\ref{app:feedback}.
A lightweight Token Selection Transformer (TST) parameterized by $\theta$ predicts a policy $\pi_\theta(a\mid s_t)$ over the action space
$\mathcal{U}(s_t)=\{\textsc{Stop}\}\cup
\{\textsc{Add}(c):c\in\mathcal{C}_t\}$.
Utility Policy Optimization (UPO) trains this policy using the retrieval effect of each action. For training instance $i$, the marginal utility of adding candidate $c$ at step $t$ is
$\Delta_{i,t}(c)
=\operatorname{AP}_i(\mathcal{S}_t\cup\{c\})
-\operatorname{AP}_i(\mathcal{S}_t)$,
where $\operatorname{AP}_i(\mathcal{S})$ denotes the retrieval AP obtained when $\mathcal{S}$ is used as the representation of sample $i$. 
We standardize this utility within retrieval direction $o$ and allocation step $t$ as
$\widetilde{R}_{i,t}(\textsc{Add}(c))
=(\Delta_{i,t}(c)-\mu_{o,t})/(\sigma_{o,t}+\epsilon)$,
where $\mu_{o,t}$ and $\sigma_{o,t}$ are the corresponding mean and standard deviation; stopping is assigned $\widetilde{R}_{i,t}(\textsc{Stop})=0$.

Let $\mathcal{D}_{\mathrm{UPO}}$ contain the allocation states generated from the training set. Since the utilities of all admissible actions are explicitly evaluated, UPO optimizes their expected retrieval utility under the predicted policy:
\begin{equation}
\mathcal{L}_{\mathrm{UPO}}
=
-\frac{1}{|\mathcal{D}_{\mathrm{UPO}}|}
\sum_{(i,t)\in\mathcal{D}_{\mathrm{UPO}}}
\sum_{a\in\mathcal{U}(s_{i,t})}
\pi_\theta(a\mid s_{i,t})
\widetilde{R}_{i,t}(a).
\label{eq:upo}
\end{equation}
Since every admissible action's marginal utility is explicitly enumerated at each step, UPO directly optimizes the expected utility under the current policy, rather than estimating it from sampled trajectories as in policy-gradient methods.

Marginal Utility Allocation (MUA) applies the learned policy to construct the final representation. Starting from the minimum valid prefix $\mathcal{S}_0$, it selects
$c_t^{\star}
=\arg\max_{c\in\mathcal{C}_t}
\pi_\theta(\textsc{Add}(c)\mid s_t)$
and denotes its probability by
$p_t^{\star}
=\pi_\theta(\textsc{Add}(c_t^{\star})\mid s_t)$.
The selected set is updated according to
\begin{equation}
\mathcal{S}_{t+1}
=
\begin{cases}
\mathcal{S}_t\cup\{c_t^{\star}\},
& p_t^{\star}>\delta,\\
\mathcal{S}_t,
& p_t^{\star}\leq\delta\quad\text{(terminate)},
\end{cases}
\label{eq:mua}
\end{equation}
where $\delta$ is the allocation threshold. After an expansion, $\mathcal{C}_{t+1}$ is updated to contain the next admissible candidates. The process terminates at step $T$ and returns $Z_x=\mathcal{S}_T$ as the CAES of sample $x$. MUA introduces no additional learnable parameters.
\begin{table*}[t]
\centering
\caption{
Full-gallery bidirectional image--text retrieval results on four datasets,
measured by mAP (\%). T2I and I2T denote text-to-image and image-to-text retrieval, respectively. Avg. Tokens is computed on COCO.
}
\label{tab:main_image_text}

\small
\setlength{\tabcolsep}{4.0pt}
\renewcommand{\arraystretch}{1.08}

\begin{tabular}{lccccccccc}
\toprule
& &
\multicolumn{2}{c}{\textbf{COCO}}
&
\multicolumn{2}{c}{\textbf{Flickr30K}}
&
\multicolumn{2}{c}{\textbf{ADE20K}}
&
\multicolumn{2}{c}{\textbf{OpenImages}} \\
\cmidrule(lr){3-4}
\cmidrule(lr){5-6}
\cmidrule(lr){7-8}
\cmidrule(lr){9-10}

\textbf{Method} & \textbf{Avg. Tokens}
& \textbf{T2I} & \textbf{I2T}
& \textbf{T2I} & \textbf{I2T}
& \textbf{T2I} & \textbf{I2T}
& \textbf{T2I} & \textbf{I2T} \\
\midrule

\rowcolor{gray!15}
\multicolumn{10}{c}{\textbf{Single-Vector Retrieval}} \\

CLIP ViT-L/14 & 1
& 9.38 & 12.78
& 15.58 & 21.42
& 7.16 & 11.80
& 3.85 & 7.71 \\

SigLIP& 1
& 23.04 & 23.16
& 32.55 & 37.20
& 15.76 & 19.87
& 9.19 & 15.65 \\

VLM2Vec-V2& 1
& 35.15 & 36.96
& 47.12 & 45.43
& 24.36 & 27.00
& 15.19 & 18.61 \\

Qwen3-VL-Embedding-2B& 1
& 41.62 & 41.11
& 56.25 & 57.05
& 28.41 & 28.63
& 14.63 & 16.36 \\

Qwen3-VL-Embedding-8B& 1
& 50.20 & 49.52
& 61.29 & 65.07
& 39.09 & 37.90
& 20.14 & 22.37 \\

\midrule
\rowcolor{gray!15}
\multicolumn{10}{c}{\textbf{Multi-Vector Retrieval (Fixed Capacity)}} \\

ColQwen2 & 578.3
& 38.95 & 24.44
& 49.26 & 30.21
& 26.84 & 16.71
& -- & -- \\

MetaEmbed (2/4)& 3
& 62.31 & 60.48
& 77.57 & 77.24
& 46.66 & 46.45
& 27.74 & 30.03 \\

MetaEmbed (4/8)& 6
& 62.33 & 60.00
& 76.97 & \textbf{77.29}
& 47.77 & \textbf{47.88}
& 29.34 & 30.93 \\

MetaEmbed (8/16)& 12
& 62.15 & 59.37
& 77.10 & 76.46
& 48.22 & 47.75
& 29.44 & 30.46 \\

MetaEmbed (16/64)& 40
& 62.27 & 59.12
& 77.10 & 76.10
& 48.22 & 47.75
& 29.33 & 30.23 \\

\midrule
\rowcolor{gray!15}
\multicolumn{10}{c}{\textbf{Multi-Vector Retrieval (Sample-Adaptive Capacity) }} \\
\textbf{AdaptiveEmbed (Ours)} & 2.1
& \textbf{63.45} & \textbf{61.14}
& \textbf{78.27} & 75.72
& \textbf{50.21} & 47.27
& \textbf{29.90} & \textbf{31.09} \\

Oracle AdaptiveEmbed & 2.2
& 69.51 & 67.85
& 82.43 & 80.88
& 57.10 & 54.59
& 37.15 & 38.29 \\

\bottomrule
\end{tabular}
\end{table*}
\section{Experiments}
\label{sec:experiments}
We evaluate SAMVR on image--text, video--text, and audio--text retrieval (Sec.~\ref{sec:main_res}), ablate AdaptiveEmbed's components (Sec.~\ref{sec:ablation}), and analyze adaptive representation and parameter sensitivity (Secs.~\ref{sec:adaptive_analysis} and~\ref{sec:sensitivity}). Appendix~\ref{app:additional} provides further experiments, including metric robustness, bank robustness, utility--cost trade-offs across modalities, allocation-behavior analyses, and examples.

\begin{table*}[t]
\centering
\caption{Generalization of sample-adaptive multi-vector representations to video--text and audio--text retrieval, measured by mAP (\%). T2V, V2T, T2A, and A2T denote text-to-video, video-to-text, text-to-audio, and audio-to-text retrieval, respectively. Avg. Tokens is computed on ActivityNet.}
\label{tab:main_video_text}

\small
\setlength{\tabcolsep}{4.0pt}
\renewcommand{\arraystretch}{1.08}

\begin{tabular}{lccccccccc}
\toprule
& &
\multicolumn{2}{c}{\textbf{ActivityNet}}
&
\multicolumn{2}{c}{\textbf{DiDeMo}}
&
\multicolumn{2}{c}{\textbf{Clotho}} 
&
\multicolumn{2}{c}{\textbf{MACS}}
\\
\cmidrule(lr){3-4}
\cmidrule(lr){5-6}
\cmidrule(lr){7-8}
\cmidrule(lr){9-10}

\textbf{Method} & \textbf{Avg. Tokens}
& \textbf{T2V} & \textbf{V2T}
& \textbf{T2V} & \textbf{V2T}
& \textbf{T2A} & \textbf{A2T}
& \textbf{T2A} & \textbf{A2T} \\
\midrule

\rowcolor{gray!15}
\multicolumn{10}{c}{\textbf{Single-Vector Retrieval}} \\
VLM2Vec-V2& 1
& 33.59 & 33.34
& 20.97 & 18.90
& -- & --
& -- & -- \\
Qwen3-VL-Embedding-2B& 1
& 60.32 & 48.46
& 47.12 & 41.42
& -- & --
& -- & -- \\

Qwen3-VL-Embedding-8B& 1
& 66.26 & 57.48
& 54.76 & 48.35
& -- & --
& -- & -- \\
LAION-CLAP& 1
& -- & --
& -- & --
& 16.21 & 20.42
& 10.21 & 9.60 \\
WAVE-7B& 1
& -- & --
& -- & --
& 30.65 & 28.47
& 10.80 & 4.75 \\
\midrule
\rowcolor{gray!15}
\multicolumn{10}{c}{\textbf{Multi-Vector Retrieval (Fixed Capacity)}} \\
MetaEmbed (2/4)& 3
& 60.86 & 52.13
& 47.26 & 44.75
& 32.03 & 30.17
& 14.93 & 10.87 \\

MetaEmbed (4/8)& 6
& 61.42 & 54.37
& 47.88 & 46.12
& 29.60 & 30.90
& 17.31 & 13.30 \\

MetaEmbed (8/16)& 12
& 61.51 & 55.60
& 47.71 & \textbf{46.62}
& 30.42 & 31.14
& 18.55 & 14.56 \\

MetaEmbed (16/64)& 40
& 60.28 & 54.34
& 46.15 & 45.10
& 29.51 & 30.24
& 18.59 & 15.12 \\

\midrule
\rowcolor{gray!15}
\multicolumn{10}{c}{\textbf{Multi-Vector Retrieval (Sample-Adaptive Capacity) }} \\
\textbf{AdaptiveEmbed (Ours)} & 1.9
& \textbf{61.62} & \textbf{56.48}
& \textbf{48.01} & 46.16
& \textbf{32.56} & \textbf{31.75}
& \textbf{20.41} & \textbf{18.01} \\
Oracle AdaptiveEmbed & 1.7
& 67.28 & 60.44
& 53.14 & 50.09
& 40.75 & 38.95
& 24.03 & 20.72 \\
\bottomrule
\end{tabular}
\end{table*}

\subsection{Experimental Setup}
\label{sec:experimental_setup}
We evaluate SAMVR on eight benchmarks spanning image--text (COCO~\citep{lin2014microsoft}, Flickr30K~\citep{young2014image},
ADE20K~\citep{zhou2017scene}, OpenImages~\citep{kuznetsova2020open}), video--text~\citep{krishna2017dense,hendricks2017localizing}, and audio--text~\citep{drossos2020clotho,martinmorato2021diversity} retrieval, unified under a full-gallery, one-to-one protocol: each query has exactly one positive item, and all baselines are re-evaluated on identical galleries and query sets. For image--text retrieval, training uses only COCO, and the remaining three datasets are evaluated strictly zero-shot; video--text and audio--text models are trained and evaluated in-domain. We report bidirectional mAP@ALL and Avg. Tokens, which denotes the mean number of active vectors, averaged over all queries in both retrieval directions. Baselines include single-vector and fixed-capacity multi-vector methods on the modalities they support. For a controlled comparison, MetaEmbed is reproduced within our framework and inherits its Matryoshka-style grouped contrastive learning: it shares the backbone, training data, and all Stage-I optimization settings with AdaptiveEmbed, differing only in the number of representation tokens and the training loss. We additionally report \emph{Oracle AdaptiveEmbed}, a non-deployable upper bound selecting the highest-utility representation per sample. Dataset statistics, protocol construction, and full implementation details are in Appendix~\ref{app:datasets}.

\begin{table*}[t]
\centering
\caption{Ablation study of AdaptiveEmbed. We evaluate the contribution of Multi-Group Contrastive Learning (MGCL) and bank feedback on bidirectional multimodal retrieval. Results are measured by mAP (\%).  Avg. Tokens is computed on COCO.}
\label{tab:ab}

\small
\setlength{\tabcolsep}{4.0pt}
\renewcommand{\arraystretch}{1.08}

\begin{tabular}{lccccccccc}
\toprule
& &
\multicolumn{2}{c}{\textbf{COCO}}
&
\multicolumn{2}{c}{\textbf{Flickr30K}}
&
\multicolumn{2}{c}{\textbf{ActivityNet}}
&
\multicolumn{2}{c}{\textbf{Clotho}} \\
\cmidrule(lr){3-4}
\cmidrule(lr){5-6}
\cmidrule(lr){7-8}
\cmidrule(lr){9-10}

\textbf{Method} & \textbf{Avg. Tokens}
& \textbf{T2I} & \textbf{I2T}
& \textbf{T2I} & \textbf{I2T}
& \textbf{T2V} & \textbf{V2T}
& \textbf{T2A} & \textbf{A2T} \\
\midrule
\textbf{AdaptiveEmbed (Ours)} & 2.1
& \textbf{63.45} & \textbf{61.14}
& \textbf{78.27} & \textbf{75.72}
& \textbf{61.62} & \textbf{56.48}
& \textbf{32.56} & \textbf{31.75} \\

w/o MGCL & 2.1
& 62.44 & 60.03
& 77.85 & 75.57
& 61.35 & 56.21
& 30.16 & 29.94 \\

w/o bank-feedback & 2.5
& 62.89 & 59.58
& 77.85 & 75.19
& 61.41 & 56.25
& 31.43 & 30.62 \\
\bottomrule
\end{tabular}
\end{table*}

\begin{figure}[t]
\centering
\includegraphics[width=\linewidth]{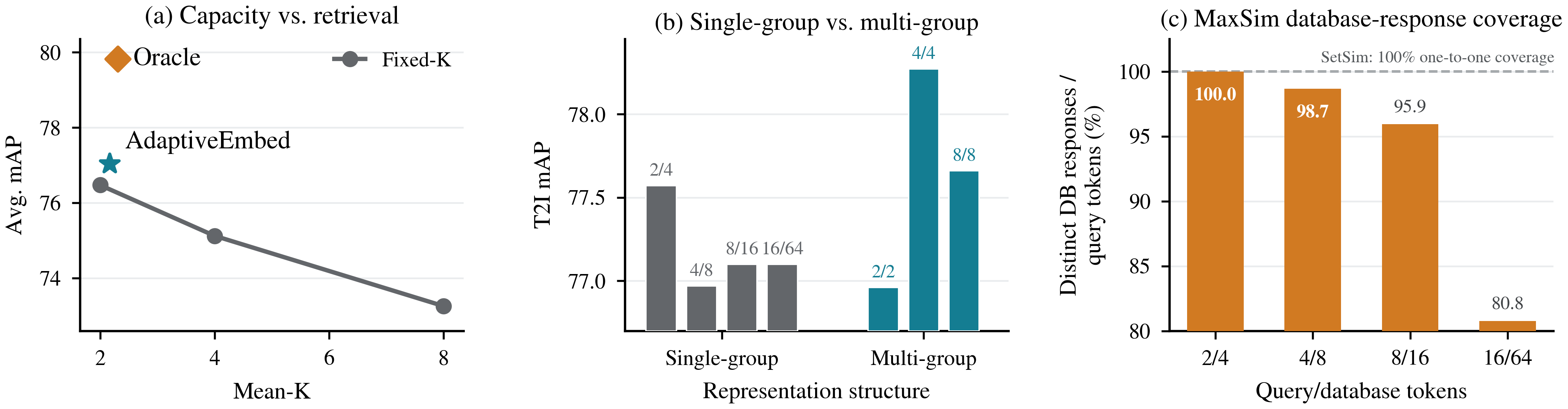}
\caption{\textbf{Analysis of adaptive representation on Flickr30K under zero-shot transfer.}
(a) Under the \emph{same} MGCL-trained representation, fixed-$K$ retrieval does not improve monotonically with capacity, whereas AdaptiveEmbed surpasses every fixed-$K$ operating point at a lower average capacity; the oracle marks the upper bound of sample-wise allocation. (b) Multi-group representation learning outperforms single-group structures under markedly different token allocations.
(c) Under MaxSim, the fraction of query tokens matched to distinct database tokens drops from $100.0\%$ to $80.8\%$ as capacity grows, whereas SetSim guarantees one-to-one coverage by construction.}
\label{fig:analysis}
\end{figure}

\subsection{Results}
\label{sec:main_res}
Table~\ref{tab:main_image_text} reports the primary image--text results. With an average of only $2.1$ tokens per query, AdaptiveEmbed attains the strongest performance on $6$ of the $8$ retrieval directions and the highest average mAP overall, both in-domain on COCO ($63.45$/$61.14$ T2I/I2T) and under strict zero-shot transfer---most notably on OpenImages, where it leads both directions against a gallery of over $546$K items. Across fixed-capacity configurations, increasing the budget from $3$ to $40$ tokens does not consistently improve retrieval, confirming that a common capacity for all samples is unnecessary. Oracle AdaptiveEmbed reaches $61.0$ average mAP at a comparable average capacity ($2.2$ tokens), showing that sample-adaptive allocation admits substantially stronger retrieval at essentially no extra representation cost---the SAMVR
setting itself carries far more potential than fixed-capacity budgets.

We next examine whether sample-adaptive allocation extends beyond image--text retrieval. Table~\ref{tab:main_video_text} reports video--text and audio--text results, keeping the same SAMVR formulation and only swapping in the modality-specific encoder. AdaptiveEmbed achieves the strongest multi-vector result on 7 of the 8 directions with an average of $1.9$ tokens on
ActivityNet, and the same pattern holds as in the image case: enlarging the fixed multi-vector budget does not consistently help, whereas per-sample allocation remains effective at a markedly smaller average representation. The oracle again shows further headroom under comparable or lower average capacity, confirming that adaptive capacity allocation is not tied to a particular
modality pair.
\subsection{Ablation Study}
\label{sec:ablation}
We ablate the two key components of AdaptiveEmbed on Table~\ref{tab:ab}:
\emph{Multi-Group Contrastive Learning (MGCL)}, which structures the underlying
representation, and \emph{MaxSim bank feedback}, which supplies retrieval-aware
observations for allocation. Removing MGCL consistently degrades performance at
a comparable budget, showing
that simply producing multiple vectors is insufficient---organizing them into
complementary groups provides a stronger basis for adaptive selection. Removing
the bank-feedback branch also lowers performance \emph{despite} a larger average
capacity,
confirming that additional vectors cannot compensate for the missing
retrieval-context signal and that feedback is an informative signal for deciding
which representation to expose per sample.

\begin{figure}[t]
  \centering
  \begin{subfigure}{0.54\textwidth}
    \centering
    \includegraphics[width=\linewidth]{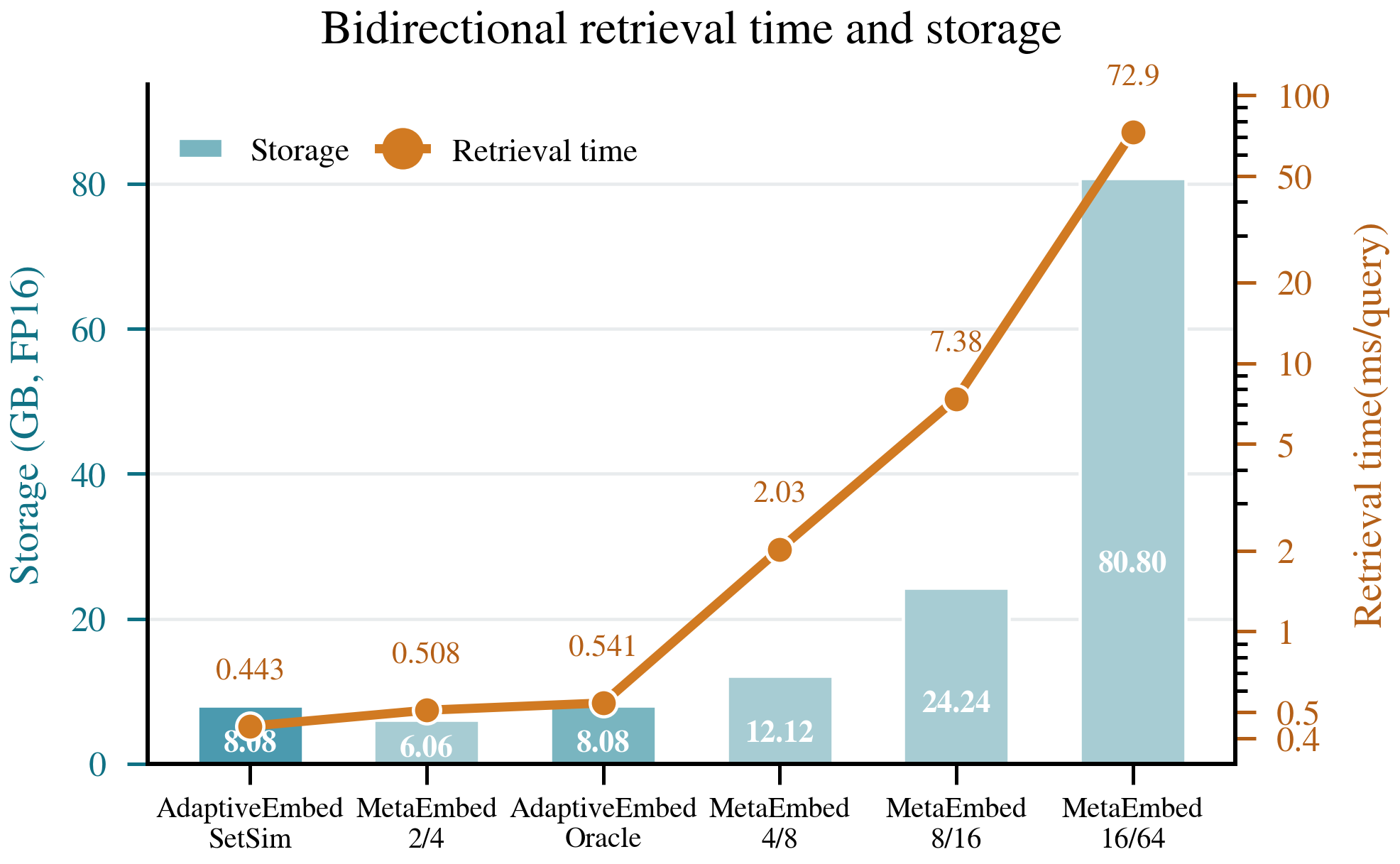}
    \caption{Bidirectional set-matching time and storage on COCO.}
    \label{fig:time}
  \end{subfigure}
  \hfill
  \begin{subfigure}{0.44\textwidth}
    \centering
    \includegraphics[width=\linewidth]{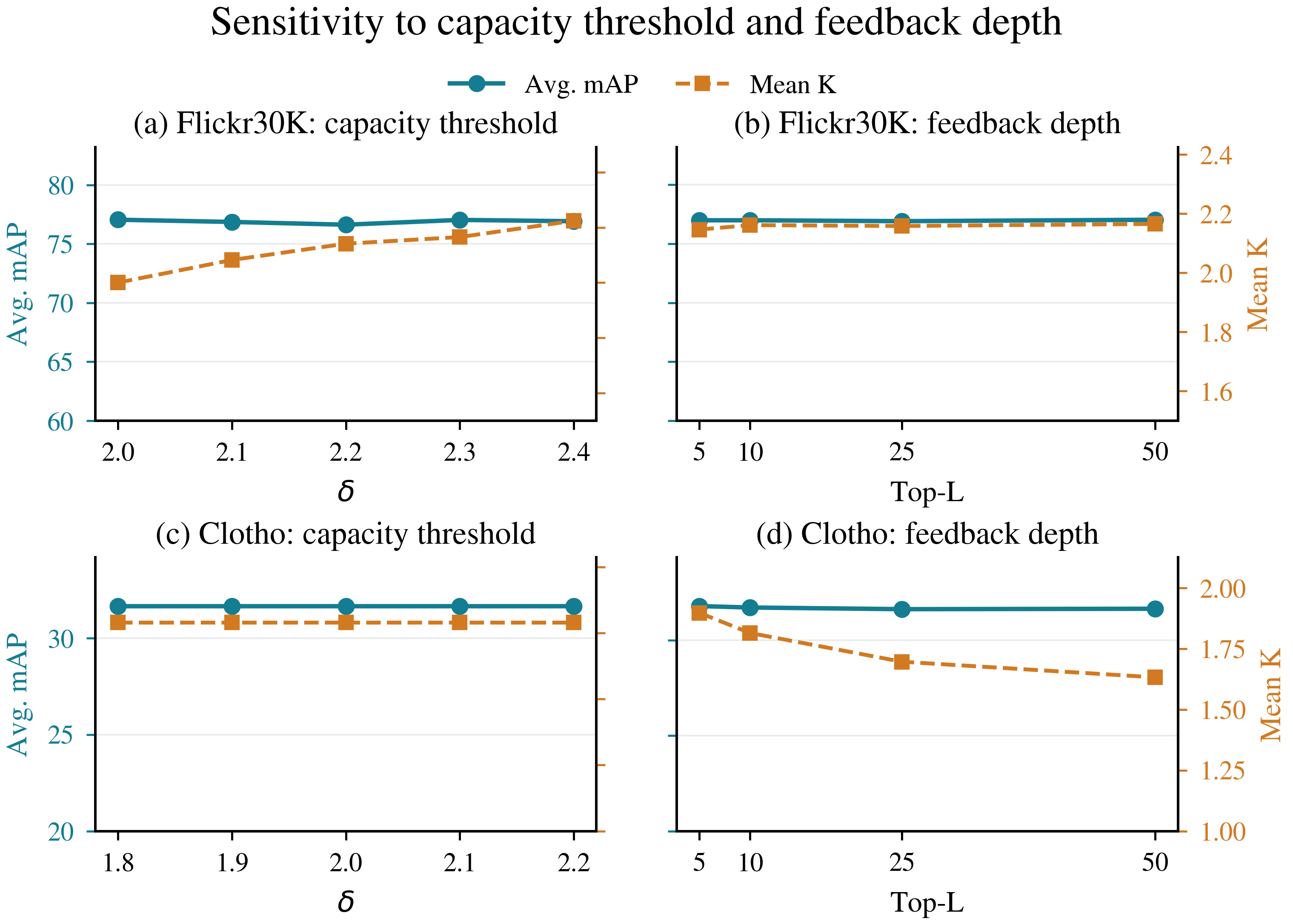}
    \caption{Sensitivity to the capacity threshold $\delta$ and feedback depth Top-$L$ on Flickr30K and Clotho.}
    \label{fig:parameter_sensitivity}
  \end{subfigure}
  \caption{Efficiency and parameter sensitivity analysis of AdaptiveEmbed.}
  \label{fig:efficiency_robustness}
\end{figure}


\subsection{Analysis of Adaptive Representation}
\label{sec:adaptive_analysis}
We analyze \emph{why} adaptive allocation works from four angles: whether the gain stems from allocation itself rather than representation capacity (Fig.~\ref{fig:analysis}(a)), whether the multi-group structure is necessary (Fig.~\ref{fig:analysis}(b)), why SetSim is the right comparison for CAES (Fig.~\ref{fig:analysis}(c)), and what adaptive allocation costs at retrieval time (Fig.~\ref{fig:time}). Appendix~\ref{app:additional} extends these analyses with additional retrieval metrics (Recall, nDCG, mean rank) on COCO and zero-shot ADE20K, bank-robustness experiments, the utility--cost trade-off and single- vs.\ multi-group comparisons across modalities, and the
decision-complexity and learned-vs.-oracle allocation distributions.

\textbf{Does adaptive allocation help beyond simply using more or fewer vectors?} Figure~\ref{fig:analysis}(a) holds the representation fixed---all operating
points share the same MGCL-trained embedding set---so only the allocation differs. Fixed-$k$ retrieval does not improve monotonically with capacity,
while AdaptiveEmbed surpasses \emph{every} fixed-$k$ point at a lower average capacity: the gain is attributable to per-sample allocation, not a stronger encoder or a larger budget. The oracle marks substantial remaining headroom.

\textbf{Does the multi-group structure matter?}
Figure~\ref{fig:analysis}(b) compares single- and multi-group structures on Flickr30K. Both follow the same coarse-to-fine prefix principle and differ only in the number of coarse anchors. Multi-group configurations match or exceed single-group ones across token allocations, indicating that distributing capacity over multiple semantic anchors, rather than deepening a single coarse-to-fine chain, provides the stronger basis for adaptive selection.

\textbf{Why SetSim rather than MaxSim?}
Figure~\ref{fig:analysis}(c) measures the fraction of query tokens whose MaxSim responses fall on distinct database tokens: coverage drops from $100.0\%$ to $80.8\%$ as MetaEmbed grows from $2/4$ to $16/64$, i.e., more query tokens collapse onto the same database tokens at larger capacity. SetSim avoids this by construction through its one-to-one assignment between equal-cardinality sets.

\textbf{What does adaptive allocation cost at retrieval time?}
Figure~\ref{fig:time} reports bidirectional retrieval time and storage on COCO. Activating only about two tokens per query, AdaptiveEmbed's set-to-set matching cost stays far below high-capacity fixed configurations, whose late-interaction cost grows sharply with the number of vectors.

\subsection{Parameter sensitivity analysis}
\label{sec:sensitivity}
We vary the capacity threshold $\delta$ and feedback depth Top-$L$ on Flickr30K and Clotho  (Fig.~\ref{fig:parameter_sensitivity}). Across the evaluated ranges, Avg.\ mAP varies by at most $0.43$ on Flickr30K and $0.16$ on Clotho, while the average capacity remains close to two tokens. Notably, increasing Top-$L$ reduces the Clotho mean capacity from $1.90$ to $1.63$ without degrading retrieval performance. These results show that AdaptiveEmbed is robust to both hyperparameters and does not depend on a narrowly tuned configuration.

\section{Conclusion}
In this work, we introduce \emph{Sample-Adaptive Multi-Vector Representation} (SAMVR), a new problem setting that explores sample-level representation capacity allocation in multi-vector representation learning for multimodal retrieval. SAMVR formulates representation capacity as a sample-dependent variable and defines \emph{content-adaptive embedding sets} (CAES) as the resulting representations, where different samples can be represented with different amounts of embedding capacity.
To investigate the feasibility of SAMVR in multimodal retrieval, we present \emph{AdaptiveEmbed}, an initial framework that combines candidate representation learning with utility-driven capacity allocation. Through Multi-Group Contrastive Learning (MGCL), set-to-set similarity (SetSim), and Utility Policy Optimization (UPO) with Marginal Utility Allocation (MUA), AdaptiveEmbed provides one possible implementation of sample-adaptive multi-vector representations.
Experiments across diverse multimodal retrieval benchmarks verify that sample-adaptive representation capacity can be learned and evaluated in practice. We hope SAMVR can serve as a starting point for further investigation into adaptive representation learning, where the allocation of representation capacity becomes an explicit research problem.

\bibliography{iclr2027_conference}

@inproceedings{xiao2026metaembed,
  title={{Metaembed: Scaling multimodal retrieval at test-time with flexible late interaction}},
  author={Xiao, Zilin and Ma, Qi and Gu, Mengting and Chen, Chun-cheng Jason and Chen, Xintao and Ordonez, Vicente and Mohan, Vijai},
  booktitle={International Conference on Learning Representations (ICLR)},
  year={2026}
}

@inproceedings{khattab2020colbert,
  title={Colbert: Efficient and effective passage search via contextualized late interaction over bert},
  author={Khattab, Omar and Zaharia, Matei},
  booktitle={Proceedings of the 43rd International ACM SIGIR Conference on Research and Development in Information Retrieval (SIGIR)},
  year={2020}
}

@inproceedings{santhanam2022colbertv2,
  title={Colbertv2: Effective and efficient retrieval via lightweight late interaction},
  author={Santhanam, Keshav and Khattab, Omar and Saad-Falcon, Jon and Potts, Christopher and Zaharia, Matei},
  booktitle={Proceedings of the 2022 Conference of the North American Chapter of the Association for Computational Linguistics (NAACL)},
  year={2022}
}

@inproceedings{pont-tuset2020localized,
  title     = {Connecting Vision and Language with Localized Narratives},
  author    = {Pont-Tuset, Jordi and Uijlings, Jasper and Changpinyo, Soravit
               and Soricut, Radu and Ferrari, Vittorio},
  booktitle = {European Conference on Computer Vision (ECCV)},
  year      = {2020}
}

@inproceedings{faysse2025colpali,
  title={Colpali: Efficient document retrieval with vision language models},
  author={Faysse, Manuel and Sibille, Hugues and Wu, Tony and Omrani, Bilel and Viaud, Gautier and Hudelot, C{\'e}line and Colombo, Pierre},
  booktitle={International Conference on Learning Representations (ICLR)},
  year={2025}
}

@inproceedings{dhulipala2024muvera,
  title={Muvera: Multi-vector retrieval via fixed dimensional encoding},
  author={Dhulipala, Laxman and Hadian, Majid and Jayaram, Rajesh and Lee, Jason and Mirrokni, Vahab},
  booktitle={Advances in Neural Information Processing Systems (NeurIPS)},
  year={2024}
}

@inproceedings{radford2021clip,
  title={Learning transferable visual models from natural language supervision},
  author={Radford, Alec and Kim, Jong Wook and Hallacy, Chris and Ramesh, Aditya and Goh, Gabriel and Agarwal, Sandhini and Sastry, Girish and Askell, Amanda and Mishkin, Pamela and Clark, Jack and others},
  booktitle={International conference on machine learning (ICML)},
  year={2021},
}

@inproceedings{zhai2023siglip,
  title={Sigmoid loss for language image pre-training},
  author={Zhai, Xiaohua and Mustafa, Basil and Kolesnikov, Alexander and Beyer, Lucas},
  booktitle={IEEE/CVF International Conference on Computer Vision (ICCV)},
  year={2023},
}

@inproceedings{li2023blip2,
  title={Blip-2: Bootstrapping language-image pre-training with frozen image encoders and large language models},
  author={Li, Junnan and Li, Dongxu and Savarese, Silvio and Hoi, Steven},
  booktitle={International conference on machine learning (ICML)},
  year={2023},
}

@article{jiang2025vlm2vec,
  title={{VLM2Vec-V2: Advancing Multimodal Embedding for Videos, Images, and Visual Documents}},
  author={Meng, Rui and Jiang, Ziyan and Liu, Ye and Su, Mingyi and Yang, Xinyi and Fu, Yuepeng and Qin, Can and Chen, Zeyuan and Xu, Ran and Xiong, Caiming and Zhou, Yingbo and Chen, Wenhu and Yavuz, Semih},
  journal={arXiv preprint arXiv:2507.04590},
  year={2025}
}

@inproceedings{martinmorato2021diversity,
  title={Diversity and bias in audio captioning datasets},
  author={Morato, Irene Martin and Mesaros, Annamaria},
  booktitle={Detection and Classication of Acoustic Scenes and Events (DCASE)},
  year={2021},
}

@article{qwen3vlembedding,
  title={Qwen3-vl-embedding and qwen3-vl-reranker: A unified framework for state-of-the-art multimodal retrieval and ranking},
  author={Li, Mingxin and Zhang, Yanzhao and Long, Dingkun and Chen, Keqin and Song, Sibo and Bai, Shuai and Yang, Zhibo and Xie, Pengjun and Yang, An and Liu, Dayiheng and others},
  journal={arXiv preprint arXiv:2601.04720},
  year={2026}
}

@inproceedings{kusupati2022matryoshka,
  title={{Matryoshka Representation Learning}},
  author={Kusupati, Aditya and Bhatt, Gantavya and Rege, Aniket and Wallingford, Matthew and Sinha, Aditya and Ramanujan, Vivek and Howard-Snyder, William and Chen, Kaifeng and Kakade, Sham and Jain, Prateek and Farhadi, Ali},
  booktitle={Advances in Neural Information Processing Systems (NeurIPS)},
  year={2022}
}

@article{oord2018cpc,
  title={Representation learning with contrastive predictive coding},
  author={Oord, Aaron van den and Li, Yazhe and Vinyals, Oriol},
  journal={arXiv preprint arXiv:1807.03748},
  year={2018}
}

@inproceedings{karpukhin2020dpr,
  title={Dense passage retrieval for open-domain question answering},
  author={Karpukhin, Vladimir and Oguz, Barlas and Min, Sewon and Lewis, Patrick and Wu, Ledell and Edunov, Sergey and Chen, Danqi and Yih, Wen-tau},
  booktitle={Proceedings of the 2020 conference on empirical methods in natural language processing (EMNLP)},
  year={2020}
}

@inproceedings{wang2021prf,
  title={Pseudo-relevance feedback for multiple representation dense retrieval},
  author={Wang, Xiao and Macdonald, Craig and Tonellotto, Nicola and Ounis, Iadh},
  booktitle={Proceedings of the 2021 ACM SIGIR International Conference on Theory of Information Retrieval (ICTIR)},
  year={2021}
}

@article{pony2026colbandit,
  title={Col-Bandit: Query-Time Top-$ K $ Estimation for Late-Interaction Retrieval},
  author={Pony, Roi and Goldfarb, Adi Raz and Naparstek, Oshri and Friedman, Idan and Barzelay, Udi and Schwartz, Eli},
  journal={arXiv preprint arXiv:2602.02827},
  year={2026}
}

@inproceedings{datta2024selective,
  title={A deep learning approach for selective relevance feedback},
  author={Datta, Suchana and Ganguly, Debasis and MacAvaney, Sean and Greene, Derek},
  booktitle={European Conference on Information Retrieval (ECIR)},
  year={2024},
}

@inproceedings{lin2014microsoft,
  title={Microsoft coco: Common objects in context},
  author={Lin, Tsung-Yi and Maire, Michael and Belongie, Serge and Hays, James and Perona, Pietro and Ramanan, Deva and Doll{\'a}r, Piotr and Zitnick, C Lawrence},
  booktitle={European conference on computer vision (ECCV)},
  year={2014},
}

@article{young2014image,
  title={From image descriptions to visual denotations: New similarity metrics for semantic inference over event descriptions},
  author={Young, Peter and Lai, Alice and Hodosh, Micah and Hockenmaier, Julia},
  journal={Transactions of the association for computational linguistics (TACL)},
  year={2014}
}

@inproceedings{zhou2017scene,
  title={Scene parsing through ade20k dataset},
  author={Zhou, Bolei and Zhao, Hang and Puig, Xavier and Fidler, Sanja and Barriuso, Adela and Torralba, Antonio},
  booktitle={IEEE conference on computer vision and pattern recognition (CVPR)},
  year={2017},
}

@article{kuznetsova2020open,
  title={The open images dataset v4: Unified image classification, object detection, and visual relationship detection at scale},
  author={Kuznetsova, Alina and Rom, Hassan and Alldrin, Neil and Uijlings, Jasper and Krasin, Ivan and Pont-Tuset, Jordi and Kamali, Shahab and Popov, Stefan and Malloci, Matteo and Kolesnikov, Alexander and others},
  journal={International journal of computer vision (IJCV)},
  year={2020},
}

@inproceedings{krishna2017dense,
  title={Dense-captioning events in videos},
  author={Krishna, Ranjay and Hata, Kenji and Ren, Frederic and Fei-Fei, Li and Carlos Niebles, Juan},
  booktitle={Proceedings of the IEEE international conference on computer vision (ICCV)},
  year={2017}
}

@inproceedings{hendricks2017localizing,
  title={Localizing moments in video with natural language},
  author={Anne Hendricks, Lisa and Wang, Oliver and Shechtman, Eli and Sivic, Josef and Darrell, Trevor and Russell, Bryan},
  booktitle={Proceedings of the IEEE international conference on computer vision (ICCV)},
  year={2017}
}

@inproceedings{drossos2020clotho,
  title={Clotho: An audio captioning dataset},
  author={Drossos, Konstantinos and Lipping, Samuel and Virtanen, Tuomas},
  booktitle={IEEE International Conference on Acoustics, Speech and Signal Processing (ICASSP)},
  year={2020},
}

@inproceedings{wu2023large,
  title={Large-scale contrastive language-audio pretraining with feature fusion and keyword-to-caption augmentation},
  author={Wu, Yusong and Chen, Ke and Zhang, Tianyu and Hui, Yuchen and Berg-Kirkpatrick, Taylor and Dubnov, Shlomo},
  booktitle={IEEE International Conference on Acoustics, Speech and Signal Processing (ICASSP)},
  year={2023},
}

@article{setsim,
author = {Kuhn, H. W.},
title = {The Hungarian method for the assignment problem},
journal = {Naval Research Logistics Quarterly},
year = {1955}
}

@inproceedings{tang2026wave,
  title={WAVE: learning unified \& versatile audio-visual embeddings with multimodal LLM},
  author={Tang, Changli and Xiao, Qinfan and Mei, Ke and Wang, Tianyi and Rao, Fengyun and Zhang, Chao},
  booktitle={International Conference on Learning Representations (ICLR)},
  year={2026}
}
\bibliographystyle{iclr2027_conference}

\appendix

\section{Detailed Formulation of Adaptive Representation Allocation}
\label{app:adaptive_allocation}
\begin{figure}[t]
\centering
\includegraphics[width=\linewidth]{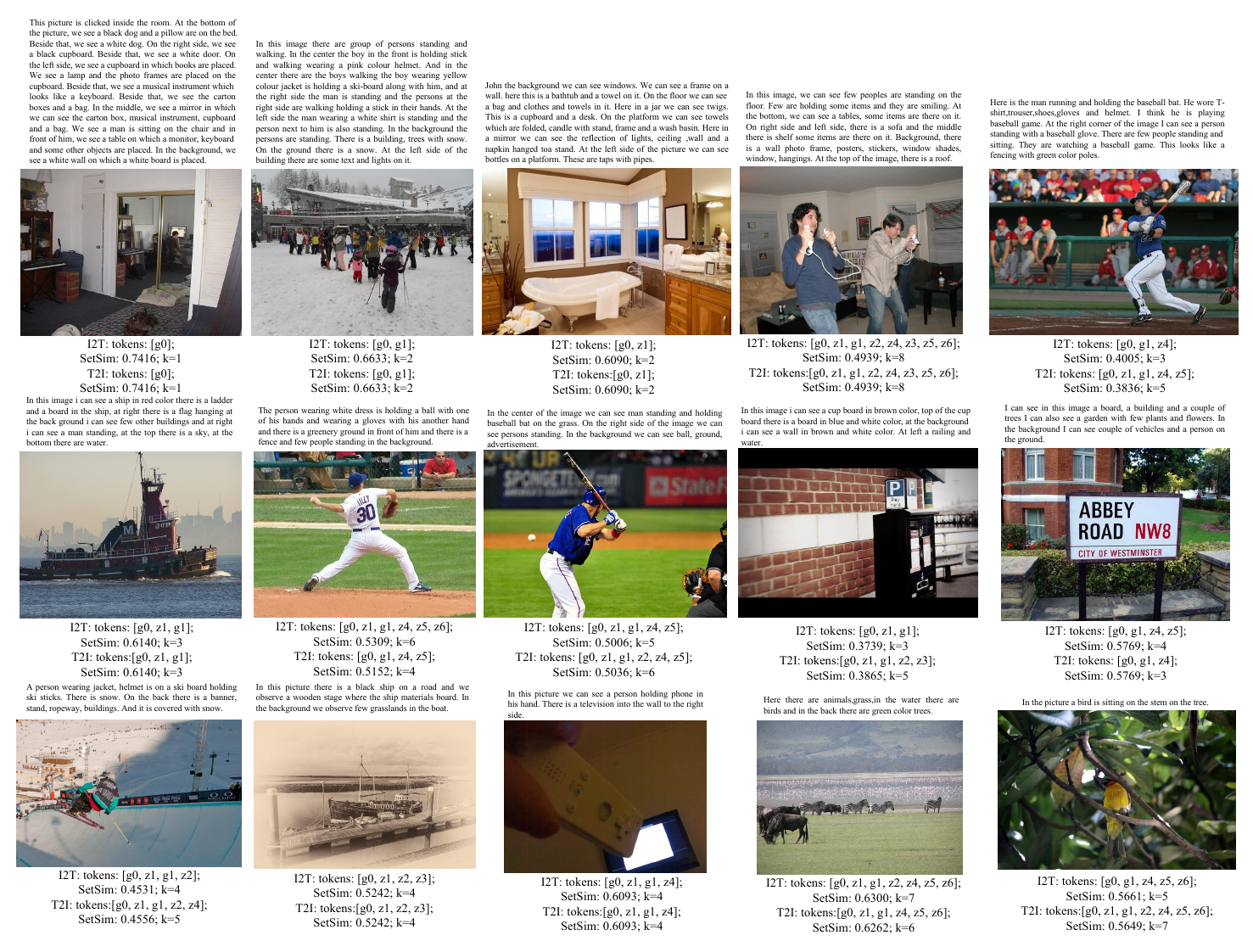}
\caption{Oracle content-adaptive embedding sets on COCO. Each sample shows
its narrative and, for both retrieval directions, the oracle-selected
tokens, the resulting SetSim score, and the capacity $k$.}
\label{fig:qualitative}
\end{figure}
This appendix provides the concrete instantiation of the sample-adaptive
allocation described formally in Sec.~\ref{sec:mgcl} and Stage~II. We specify
the group configuration, the multi-step decision process used by AdaptiveEmbed,
the construction of the bank feedback vector, and the oracle upper bound.

\subsection{Group Configuration}
\label{app:group_config}

In our instantiation, the candidate embedding set $E_x$ contains $N=8$ tokens,
organized by MGCL into $G=2$ ordered groups of equal size:
\begin{equation}
\mathcal{G}^{(1)}=\{g_0,z_1,z_2,z_3\},
\qquad
\mathcal{G}^{(2)}=\{g_1,z_4,z_5,z_6\}.
\end{equation}
Here $g_0$ and $g_1$ denote the coarse (primary) tokens that carry global
semantics, while $z_1,\ldots,z_6$ denote fine (complementary) tokens that
provide additional discriminative detail. Each group follows an ordered prefix
structure, so that shorter prefixes yield compact representations and longer
prefixes progressively add fine-grained information. The two coarse tokens
$g_0,g_1$ head the two groups respectively, making the group boundary aligned
with the coarse-to-fine transition.

\subsection{Multi-Step Adaptive Decision}
\label{app:decision}

Although the group configuration admits many prefix combinations,
AdaptiveEmbed does not predict a representation from the full combinatorial
space at once. Instead, it constructs the CAES through a short sequence of
decisions, each expanding the current representation along an admissible group
prefix or terminating with \textsc{Stop}. Our instantiation uses a three-step
process, and every step may terminate early.

Starting from the minimal representation $\mathcal{S}_0=\{g_0\}$, the decisions
are:
\begin{align}
\text{Step 1:}\quad
\{g_0\}\;\rightarrow\;&
\big\{\,\{g_0,g_1\},\;\{g_0,z_1\},\;\textsc{Stop}\,\big\},\\
\text{Step 2:}\quad
\{g_0,g_1\}\ \text{or}\ \{g_0,z_1\}\;\rightarrow\;&
\big\{\,\{g_0,g_1,z_1,z_4\},\;\textsc{Stop}\,\big\},\\
\text{Step 3:}\quad
\{g_0,g_1,z_1,z_4\}\;\rightarrow\;&
\big\{\,\{g_0,z_1,z_2,z_3,g_1,z_4,z_5,z_6\},\;\textsc{Stop}\,\big\}.
\end{align}
The first decision determines whether the coarse token $g_0$ already suffices
(\textsc{Stop}), whether a second coarse token should be added
($\to\{g_0,g_1\}$), or whether a fine token should be introduced
($\to\{g_0,z_1\}$). If expansion continues, the second decision merges toward
the balanced four-token representation $\{g_0,g_1,z_1,z_4\}$, and the third
decision optionally completes the full eight-token representation. Each choice
corresponds to one \textsc{Add}/\textsc{Stop} action in Stage~II
(Eq.~\ref{eq:mua}), and allocation terminates as soon as the predicted
expansion probability falls below the threshold $\delta$.

\paragraph{Reachable representations.}
Because each step may stop, the policy reaches one of five terminal CAES
configurations, spanning capacities from a single token to the full set:
\begin{center}
\renewcommand{\arraystretch}{1.2}
\begin{tabular}{cll}
\toprule
Capacity & Terminal CAES & Terminating decision \\
\midrule
$1$ & $\{g_0\}$ & \textsc{Stop} at Step~1 \\
$2$ & $\{g_0,g_1\}$ & expand-coarse, \textsc{Stop} at Step~2 \\
$2$ & $\{g_0,z_1\}$ & expand-fine, \textsc{Stop} at Step~2 \\
$4$ & $\{g_0,g_1,z_1,z_4\}$ & \textsc{Stop} at Step~3 \\
$8$ & $\{g_0,z_1,z_2,z_3,g_1,z_4,z_5,z_6\}$ & expand at Step~3 \\
\bottomrule
\end{tabular}
\end{center}
This hierarchical process allocates capacity according to sample-specific
retrieval utility rather than assigning a fixed number of tokens to all
samples.

\paragraph{Example trajectories.}
The five reachable configurations correspond to the following decision
trajectories:
\begin{itemize}
\item \emph{Coarse-sufficient.} The initial coarse token already discriminates
the sample:
\[
\{g_0\}\;\xrightarrow{\textsc{Stop}}\;\{g_0\}.
\]
\item \emph{Coarse-expanded.} A second coarse token resolves residual global
ambiguity, without fine detail:
\[
\{g_0\}\;\xrightarrow{\text{expand-coarse}}\;\{g_0,g_1\}
\;\xrightarrow{\textsc{Stop}}\;\{g_0,g_1\}.
\]
\item \emph{Fine-refined.} A small amount of complementary detail is added on top
of the single coarse token:
\[
\{g_0\}\;\xrightarrow{\text{expand-fine}}\;\{g_0,z_1\}
\;\xrightarrow{\textsc{Stop}}\;\{g_0,z_1\}.
\]
\item \emph{Balanced.} Both groups are partially activated to a balanced
four-token representation:
\[
\{g_0\}\;\rightarrow\;\{g_0,g_1\}\ \text{or}\ \{g_0,z_1\}
\;\rightarrow\;\{g_0,g_1,z_1,z_4\}
\;\xrightarrow{\textsc{Stop}}\;\{g_0,g_1,z_1,z_4\}.
\]
\item \emph{Full-capacity.} Rich or ambiguous content triggers expansion to the
complete eight-token set:
\[
\{g_0\}\;\rightarrow\;\{g_0,g_1,z_1,z_4\}
\;\rightarrow\;\{g_0,z_1,z_2,z_3,g_1,z_4,z_5,z_6\}.
\]
\end{itemize}

\subsection{Oracle Upper Bound}
\label{app:oracle}

The Oracle AdaptiveEmbed reported in the main tables is a non-deployable upper
bound that measures the potential of sample-adaptive allocation. It follows the
\emph{same} progressive expansion principle as the deployed policy
(Appendix~\ref{app:decision})---starting from $\{g_0\}$ and repeatedly deciding
whether to extend one of the two ordered groups by its next prefix element or to
terminate---but at a finer granularity. Whereas the deployed policy expands in
coarse blocks and merges branches, terminating in three steps at five terminal
configurations, the oracle expands one token at a time and keeps all branches
distinct, so its trajectories may take a variable number of steps and reach
\emph{every} admissible configuration rather than a merged subset.

Concretely, we take the ordered prefixes of each group,

\begin{equation}
\begin{aligned}
\mathcal{P}^{(1)}&=\{\{g_0\},\{g_0,z_1\},\{g_0,z_1,z_2\},\{g_0,z_1,z_2,z_3\}\},\\
\mathcal{P}^{(2)}&=\{\emptyset,\{g_1\},\{g_1,z_4\},\{g_1,z_4,z_5\},\{g_1,z_4,z_5,z_6\}\},
\end{aligned}
\end{equation}
and form all combinations of one prefix from each group, which yields the finite
set of reachable configurations
\begin{equation}
\mathcal{M}=\{\,p^{(1)}\cup p^{(2)} \mid p^{(1)}\in\mathcal{P}^{(1)},\,
p^{(2)}\in\mathcal{P}^{(2)}\,\},
\qquad |\mathcal{M}|=20 ,
\end{equation}
i.e., the full Cartesian product of the two groups' prefixes ($g_0$ is always
active). Representative trajectories include:
\begin{align*}
\{g_0\}\;&\xrightarrow{\textsc{Stop}}\;\{g_0\},\\
\{g_0\}\;&\rightarrow\;\{g_0,z_1\}\;\xrightarrow{\textsc{Stop}}\;\{g_0,z_1\},\\
\{g_0\}\;&\rightarrow\;\{g_0,z_1\}\;\rightarrow\;\{g_0,z_1,z_2\}
\;\rightarrow\;\{g_0,z_1,z_2,z_3\}\;\xrightarrow{\textsc{Stop}}\;
\{g_0,z_1,z_2,z_3\},\\
\{g_0\}\;&\rightarrow\;\{g_0,g_1\}\;\rightarrow\;\{g_0,g_1,z_1\}
\;\rightarrow\;\cdots\;\rightarrow\;\{g_0,z_1,z_2,z_3,g_1,z_4,z_5,z_6\},\\
&\;\;\vdots
\end{align*}
The remaining trajectories differ only in the order and depth to which each
group is expanded, and all of them terminate in one of the $20$ configurations
of $\mathcal{M}$.

For each sample $x$, the oracle scores every reachable configuration under the
retrieval objective and selects the best one:
\begin{equation}
\mathcal{Z}^{\star}(x)=\arg\max_{\mathcal{Z}\in\mathcal{M}} \operatorname{AP}(x,\mathcal{Z}),
\end{equation}
where $\operatorname{AP}(x,\mathcal{Z})$ denotes the retrieval AP obtained when $\mathcal{Z}$ is used as the representation of $x$, consistent with $\operatorname{AP}_i$ in Stage~II.
In implementation, we evaluate these $20$ configurations directly, since trajectories reaching the same configuration are equivalent under the retrieval objective. The oracle does not correspond to a deployable model, as it requires per-sample exhaustive search under the target metric; it serves only to quantify how closely the learned policy approaches the best achievable sample-wise allocation. The gap between AdaptiveEmbed and this oracle in Tables~\ref{tab:main_image_text} and~\ref{tab:main_video_text} therefore separates the potential of the SAMVR setting from the capability of the current allocation model.

\subsection{Bank Feedback Construction}
\label{app:feedback}

The feedback matrix $\mathcal{F}_t$ used in Stage~II lets the capacity policy
observe not only the current sample's tokens but also how those tokens actually
respond against a retrieval bank. If a candidate token yields a response pattern
that differs from the already-selected tokens on the current top-scoring bank
items, it is more likely to carry incremental retrieval information; otherwise
the policy can safely stop expanding.

\paragraph{Construction.}
Let the L2-normalized query tokens be $Q=\{q_i\}_{i=1}^{N}$ and the tokens of the $n$-th bank item be $D_n=\{d_{n,j}\}_{j=1}^{R_d}$. For each query token we compute its MaxSim response to bank item $n$:
\begin{equation} A_{i,n}=\max_{1\leq j\leq R_d} q_i^\top d_{n,j}.
\end{equation}
At decision state $t$ with selected token set $\mathcal{S}_t$, the current
retrieval score of bank item $n$ is
\begin{equation}
s_n^{(t)}=\frac{1}{|\mathcal{S}_t|}\sum_{i\in\mathcal{S}_t} A_{i,n},
\end{equation}
and we retain the Top-$L$ bank items under this score,
$\mathcal{N}_t=\operatorname{TopL}_{n}\big(s_n^{(t)}\big)$. The feedback matrix
collects the responses of \emph{all} tokens (selected and candidate) to these
items:
\begin{equation}
F_t=\big[A_{i,n}\big]_{\,1\leq i\leq N,\; n\in\mathcal{N}_t}\in\mathbb{R}^{N\times L}.
\end{equation}
Because $\mathcal{N}_t$ is derived only from the current retrieval score
$s_n^{(t)}$, the policy sees how candidate tokens would respond to the items that
currently matter most, without any supervision signal. In our implementation the
cached feedback tensor has shape
$[\,\text{batch},\ \text{decision context},\ 8\ \text{tokens},\ \text{Top-}L\,]$.

\paragraph{Encoding.}
\label{tst}
In the following, $H_{\mathcal{S}_t}$ and $H_{\mathcal{C}_t}$ denote the hidden representations of the selected and candidate tokens produced by the TST, $m_{\mathcal{S}_t}$ and $m_{\mathcal{C}_t}$ are binary masks indicating selected and candidate positions, $\operatorname{Norm}_{\mathcal{S}_t}(\cdot)$ standardizes each token's feedback by the response statistics of the selected set, $\operatorname{MHA}(q,k,v)$ is multi-head attention, $\operatorname{LN}$ is layer normalization, and $e_t$ is a learnable decision-context embedding for step $t$. The policy first lets candidate tokens attend to the selected tokens:
\begin{equation}
H_t=\operatorname{LN}\!\big(H_{\mathcal{C}_t}
+\operatorname{MHA}(H_{\mathcal{C}_t},H_{\mathcal{S}_t},H_{\mathcal{S}_t})\big).
\end{equation}
The feedback matrix is normalized by the response statistics of the selected
tokens and concatenated with selected/candidate masks before encoding:
\begin{equation}
E_t=\operatorname{MLP}\!\big(\big[\operatorname{Norm}_{\mathcal{S}_t}(F_t^\top);\,
m_{\mathcal{S}_t};\,m_{\mathcal{C}_t}\big]\big).
\end{equation}
Candidate tokens then read the encoded database feedback, and a pooled
representation combined with a decision-context embedding $e_t$ produces the
action logits:
\begin{align}
\widetilde{H}_t&=\operatorname{LN}\!\big(H_t+\operatorname{MHA}(H_t,E_t,E_t)\big),\\
\ell_t&=\operatorname{Head}\!\big(\operatorname{Pool}(\widetilde{H}_t)+e_t\big),
\end{align}
where $\ell_t$ are the logits over \textsc{Stop}, expansion, and branch choices. The Top-$L$ selection depends solely on the current retrieval score $s_n^{(t)}$;
the policy never observes ground-truth identities, positive/negative labels,
retrieval ranks, or AP values. The standardized marginal utilities
$\widetilde{R}_{i,t}$ used in UPO (Eq.~\ref{eq:upo}) are computed with detached
retrieval metrics and serve only as optimization targets, strictly separated
from the feedback that the policy takes as input.

\begin{table*}[t]
\centering
\small
\setlength{\tabcolsep}{3pt}
\caption{Dataset statistics and unified full-gallery retrieval protocols.
The Train column denotes the gallery prefix used for representation or
policy training when applicable. Each query has exactly one positive
item in the full gallery.}
\label{tab:dataset_statistics}
\begin{tabular}{llrrrrl}
\toprule
Modality & Dataset
& Train & Queries & Gallery
& Query split & Text construction \\
\midrule
Image--Text
& COCO       & 118,287 & 5,000  & 123,287 & val
& Full localized narrative \\
& Flickr30K  & 29,783  & 1,000  & 30,783  & val
& Full localized narrative \\
& ADE20K     & 20,210  & 2,000  & 22,210  & val
& Full localized narrative \\
& OpenImages & 504,413 & 41,620 & 546,033 & val
& Full localized narrative \\
\midrule
Video--Text
& ActivityNet   & 10,009 & 4,917 & 14,926 & val1
& Temporally ordered event captions \\
& DiDeMo        & 8,395  & 1,004 & 9,399  & test
& Joined moment descriptions \\
\midrule
Audio--Text
& MACS          & 3,537 & 393   & 3,930 & test
& Joined 2--5 captions \\
& Clotho        & 3,839 & 1,045 & 4,884 & evaluation
& Joined five captions \\
\bottomrule
\end{tabular}
\end{table*}
\section{Datasets and Evaluation Protocols}
\label{app:datasets}

\paragraph{Unified full-gallery retrieval.}
We evaluate AdaptiveEmbed on eight datasets spanning image--text, video--text,
and audio--text retrieval. To make the protocols comparable across modalities,
we convert each dataset into a one-to-one retrieval benchmark: each media item
is paired with a single complete textual description, and each query
has exactly one positive item in the gallery. The text side is constructed as follows. For image--text retrieval, we use the official Localized Narratives
annotations~\citep{pont-tuset2020localized}, which provide one long-form
narrative per image and cover exactly our four image datasets (COCO,
Flickr30K, ADE20K, and Open Images). For video--text and audio--text
retrieval, each sample's original captions are concatenated in their original order into a single long text, so that caption count does not change the number of positives across samples. The same
text construction is applied to every baseline. Under this protocol, a query
$q$ whose positive item has rank $r_q$ satisfies $\mathrm{AP}(q)=1/r_q$, so
mAP and MRR coincide; we report mAP. All baselines are re-evaluated with
identical query sets, galleries, text construction, and positive mappings. We adopt Localized Narratives rather than the original short captions
deliberately: under the standard 5-caption protocols, strong single-vector
embeddings already approach saturation, leaving little room to differentiate
multi-vector designs, whereas long-form narratives contain multiple facets
per sample and thus stress exactly the fine-grained capacity that
multi-vector representations are meant to provide.

\paragraph{Training and zero-shot splits.}
The representation learner and capacity policy are trained only on the source-domain split of each modality: COCO for image--text, the joint ActivityNet+DiDeMo pairs for video, and the joint MACS+Clotho pairs for audio. Flickr30K, ADE20K, and OpenImages are evaluated strictly zero-shot. Per-dataset statistics and splits are summarized in
Table~\ref{tab:dataset_statistics}.

\paragraph{Baselines.}
We compare SAMVR with representative single-vector and fixed-capacity multi-vector retrieval methods.
Single-vector baselines include CLIP~\cite{radford2021clip}, SigLIP~\cite{zhai2023siglip}, VLM2Vec-V2~\cite{jiang2025vlm2vec}, Qwen3-VL-Embedding~\cite{qwen3vlembedding}, LAION-CLAP~\cite{wu2023large}, and WAVE-7B~\cite{tang2026wave} when applicable to the corresponding modalities.
Multi-vector baselines include ColQwen2~\cite{faysse2025colpali} and MetaEmbed~\cite{xiao2026metaembed}.
For MetaEmbed $(R_q/R_d)$, $R_q$ and $R_d$ represent the numbers of query- and database-side embedding vectors, respectively.
We additionally report Oracle AdaptiveEmbed as a non-deployable upper bound, which selects representations according to target retrieval utility and is only used to analyze the potential of sample-adaptive capacity allocation.

\paragraph{Implementation Details.}
\label{app:implementation}
We use frozen Qwen3-VL-Embedding-2B encoders for image--text and video--text retrieval, and a frozen WAVE-7B encoder for audio--text retrieval. The Q-Former is initialized from scratch with four layers, eight attention heads, and a hidden dimension of $512$. Stage~I produces eight representation tokens per modality---one frozen global token and seven learned tokens---forming two four-token groups $\{g_0,z_1,z_2,z_3\}$ and $\{g_1,z_4,z_5,z_6\}$, and is optimized with MGCL for $20$ epochs using AdamW (learning rate $1\times10^{-4}$, effective batch size $2{,}048$, temperature $\tau=0.03$). In Stage~II, the TST uses two four-head attention blocks with hidden dimension $128$ and is trained with UPO for $20$ epochs, using the top-$50$ responses from the full source-domain training bank as feedback. Relative utility rewards are normalized independently for each retrieval direction and allocation stage, and the direction-specific MUA threshold $\delta$ is selected on source-domain validation data and fixed at test time. SetSim is computed with a batched dynamic program over token subsets, which finds the exact optimal assignment for $k\le 8$ tokens efficiently across all query--gallery pairs in a batch.

\section{Additional Experiments and Analyses}
\label{app:additional}
\subsection{Robustness Across Retrieval Metrics}
\label{app:metrics}
The main tables report mAP under the unified full-gallery protocol. To verify that the conclusions are not specific to this metric, we additionally evaluate Recall@$K$, nDCG, and mean rank on COCO (in-domain) and ADE20K (zero-shot), using the same query sets, galleries, and controlled MetaEmbed reproduction as in Sec.~\ref{sec:experimental_setup}. All numbers are averaged over both retrieval directions; $\bar{k}$ denotes the average number of activated tokens per query.

\begin{table}[t]
\centering
\small
\setlength{\tabcolsep}{5.0pt}
\renewcommand{\arraystretch}{1.08}
\caption{Additional retrieval metrics on COCO under the unified full-gallery protocol ($5{,}000$ queries, $123{,}287$ gallery items), averaged over both directions. AdaptiveEmbed leads every Recall and nDCG metric at the lowest average capacity, indicating that the conclusions of Table~\ref{tab:main_image_text} are not an artifact of the mAP metric.}
\label{tab:coco_metrics}
\begin{tabular}{lccccccc}
\toprule
\textbf{Method} & $\bar{k}$ & \textbf{R@1} & \textbf{R@5} & \textbf{R@10}
& \textbf{nDCG} & \textbf{mAP} & \textbf{MeanR}$\downarrow$ \\
\midrule
MetaEmbed (2/4)   & 3  & 51.79 & 72.54 & 79.16 & 69.13 & 61.40 & 30.58 \\
MetaEmbed (4/8)   & 6  & 51.52 & 72.10 & 79.30 & 68.96 & 61.17 & 31.34 \\
MetaEmbed (8/16)  & 12 & 50.99 & 71.86 & 79.05 & 68.65 & 60.76 & \textbf{30.18} \\
MetaEmbed (16/64) & 40 & 50.92 & 71.88 & 78.78 & 68.59 & 60.69 & 30.78 \\
\textbf{AdaptiveEmbed (Ours)} & \textbf{2.1}
& \textbf{52.86} & \textbf{73.32} & \textbf{79.55} & \textbf{69.85}
& \textbf{62.30} & 31.26 \\
\bottomrule
\end{tabular}
\end{table}
\begin{table}[t]
\centering
\small
\setlength{\tabcolsep}{5.0pt}
\renewcommand{\arraystretch}{1.08}
\caption{Additional retrieval metrics on ADE20K under strict zero-shot transfer ($2{,}000$ queries, $22{,}210$ gallery items), averaged over both directions. The representation, policy, threshold, and retrieval bank are all taken from COCO without any ADE20K tuning. AdaptiveEmbed outperforms every
MetaEmbed configuration on all Recall, nDCG, and mAP metrics at the lowest average capacity.}
\label{tab:ade_metrics}
\begin{tabular}{lccccccc}
\toprule
\textbf{Method} & $\bar{k}$ & \textbf{R@1} & \textbf{R@5} & \textbf{R@10}
& \textbf{nDCG} & \textbf{mAP} & \textbf{MeanR}$\downarrow$ \\
\midrule
MetaEmbed (2/4)   & 3  & 35.83 & 58.40 & 67.75 & 56.81 & 46.56 & 46.09 \\
MetaEmbed (4/8)   & 6  & 37.28 & 59.78 & 68.60 & 57.88 & 47.83 & \textbf{44.26} \\
MetaEmbed (8/16)  & 12 & 37.33 & 59.88 & 69.25 & 58.03 & 47.99 & 44.49 \\
MetaEmbed (16/64) & 40 & 37.28 & 59.53 & 68.93 & 57.84 & 47.81 & 48.08 \\
\textbf{AdaptiveEmbed (Ours)} & \textbf{2.0}
& \textbf{38.08} & \textbf{61.33} & \textbf{69.78} & \textbf{58.66}
& \textbf{48.74} & 46.04 \\
\bottomrule
\end{tabular}
\end{table}
On COCO, as reported in Table~\ref{tab:coco_metrics}, AdaptiveEmbed leads every Recall and nDCG metric against all MetaEmbed configurations while
activating the fewest tokens on average. The margins are largest at the top
of the ranking, with a bidirectional R@1 gain of $1.1$ points and T2I R@1
gains of $1.1$ to $1.5$ points depending on the configuration. The single
exception is mean rank, where the 8/16 configuration is marginally better,
at $30.2$ versus $31.3$, indicating that a small number of tail queries rank
slightly lower under adaptive allocation; since all top-$K$ and
ranking-quality metrics favor AdaptiveEmbed, the main conclusions are
unaffected.

The zero-shot ADE20K results in Table~\ref{tab:ade_metrics} are stronger
still. Against the capacity-matched 2/4 configuration, AdaptiveEmbed
improves R@1, R@5, and R@10 by $2.25$, $2.93$, $2.03$, and $1.38$
points respectively, and mAP by $2.18$ points. Even against the per-metric
best among all four fixed configurations, it remains ahead on every Recall,
nDCG, and mAP metric, including gains of $0.75$ points on both R@1 and mAP.
These improvements are driven primarily by the T2I direction, in which
AdaptiveEmbed leads on all metrics including mean rank, at $43.4$ versus
$45.7$, whereas I2T is mixed at the level of individual metrics; the T2I
advantage is large enough that every bidirectional average favors adaptive
allocation. As on COCO, mean rank is the one aggregate exception, at $46.0$
versus $44.3$ for the 4/8 configuration, again reflecting a small number of
poorly ranked tail queries rather than a degradation in overall retrieval
quality. Notably, this margin is achieved under strict zero-shot transfer
with no ADE20K-specific tuning of any component, indicating that the learned
allocation policy generalizes across domains rather than exploiting
source-domain particulars.

\paragraph{Retrieval Bank Composition and Robustness.}
\label{app:bank}
The retrieval bank $\mathcal{R}$ used for bank feedback
(Appendix~\ref{app:feedback}) is drawn from the source-domain training split of
each modality: the COCO training split for image--text, the joint
ActivityNet+DiDeMo training pairs for video--text, and the joint MACS+Clotho
training pairs for audio--text. The same bank is used for both UPO training and
test-time allocation; in particular, all zero-shot image--text evaluations
(Flickr30K, ADE20K, OpenImages) reuse the COCO bank, so no target-domain data
enters the feedback path.

To assess sensitivity to the bank domain, we run a bank-replacement experiment
on Flickr30K: keeping the trained representation and policy fixed, we swap the
COCO bank for a bank built from the Flickr30K training split. The in-domain
bank changes Avg.\ mAP by only $+0.07$, indicating that the allocation policy
is largely insensitive to the domain of the bank. This is consistent with the
role the bank plays in our design: it serves as a generic response
environment---a population of items against which the distinctiveness of
selected and candidate tokens is measured---rather than a source of
domain-specific retrieval content. As long as the bank is sufficiently
diverse, the marginal-utility signal it induces transfers across domains, and
matching the bank to the target domain yields no meaningful additional
benefit.
\begin{figure}[t]
\centering
\includegraphics[width=\linewidth]{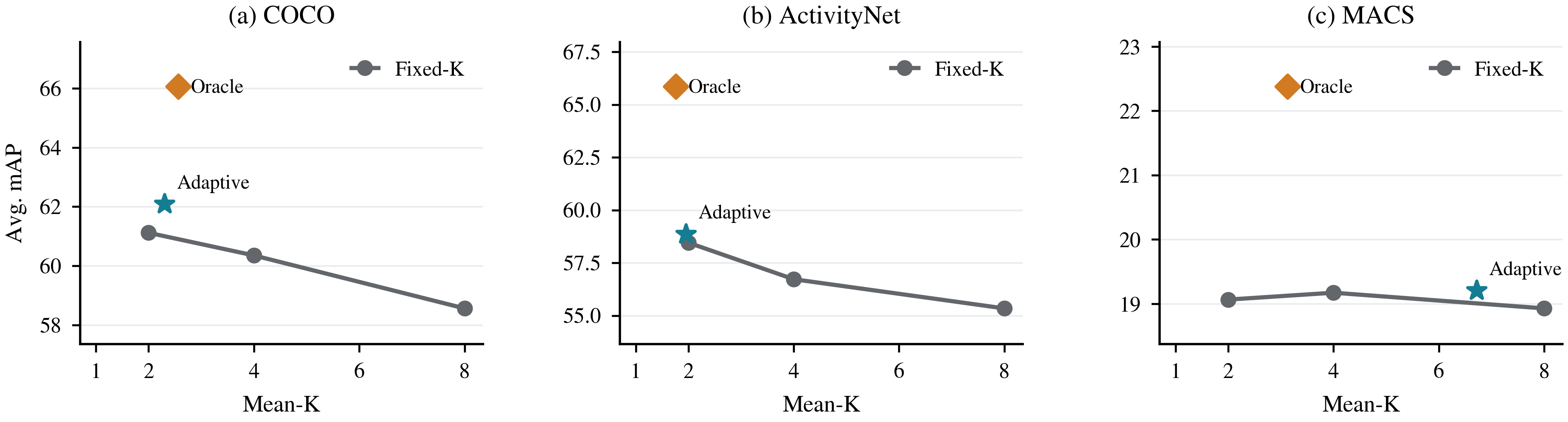}
\caption{
Retrieval utility versus representation cost under adaptive and fixed-capacity
allocation, all using the same MGCL-trained representation.
}
\label{fig:adaptive_tradeoff}
\end{figure}
\begin{figure}[t]
\centering
\includegraphics[width=\linewidth]{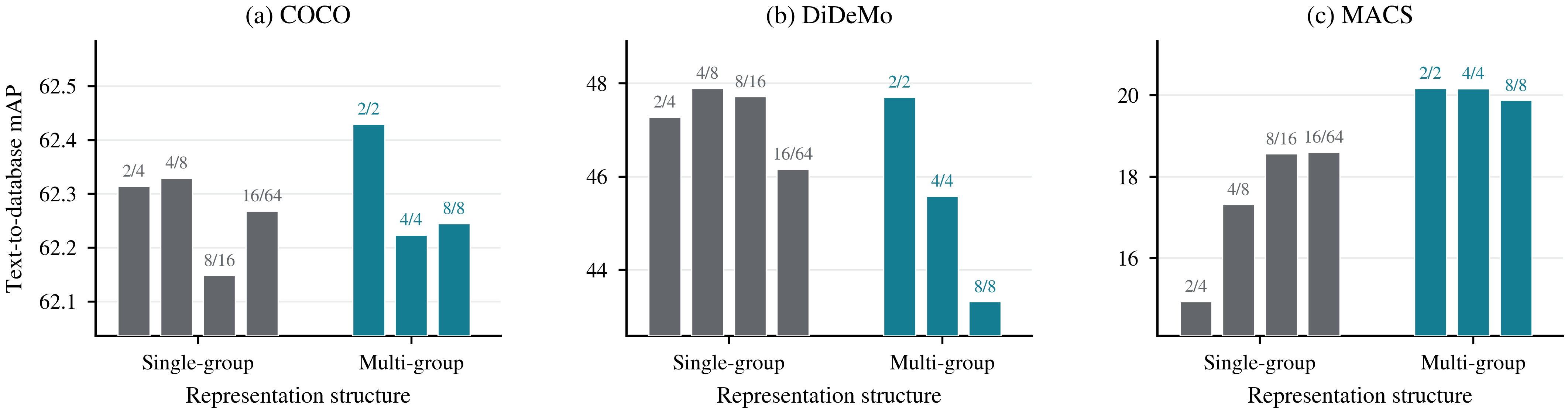}
\caption{
Ablation of single-group and multi-group representation structures.
Multi-group contrastive learning consistently improves retrieval performance
across different datasets, demonstrating that separating representation
capacity into multiple semantic groups provides a more effective basis for
adaptive token activation.
}
\label{fig:mgcl_ablation}
\end{figure}
\paragraph{Utility--Cost Trade-off Across Modalities.}
\label{app:tradeoff}
Figure~\ref{fig:adaptive_tradeoff} extends the Flickr30K analysis of
Fig.~\ref{fig:analysis}(a) to COCO, ActivityNet, and MACS, plotting average
mAP against mean activated capacity for fixed-$k$ operating points, the
learned policy, and the oracle, all sharing the same MGCL-trained
representation. The same qualitative pattern holds across all three
modalities: fixed-capacity retrieval does not improve---and often
degrades---as the shared budget grows, while AdaptiveEmbed sits above the
fixed-capacity frontier at roughly two activated tokens, and the oracle marks
substantial further headroom at a comparable or lower average capacity. The
consistency across image, video, and audio indicates that the utility--cost
advantage of per-sample allocation is a property of the formulation rather
than of a particular modality pair.

\paragraph{Single- vs.\ Multi-Group Structures Across Datasets.}
\label{app:group_structure}
Figure~\ref{fig:mgcl_ablation} complements Fig.~\ref{fig:analysis}(b) by
comparing single-group and multi-group representation structures on COCO,
DiDeMo, and MACS under matched token allocations. Both structures follow the
same coarse-to-fine prefix principle, in which the leading token carries
global semantics and subsequent tokens add complementary detail; they differ
in how many such coarse anchors the representation contains. The single-group
structure places all fine tokens behind one coarse token, whereas the
multi-group structure maintains two coarse anchors, each refined by its own
fine tokens. Multi-group configurations match or outperform their
single-group counterparts on all three datasets, with the largest gains on
MACS, where the audio--text pairs are shortest and a single global anchor is
most easily saturated. Together with the component ablation in
Table~\ref{tab:ab}, this indicates that the benefit stems from distributing
capacity across multiple semantic anchors rather than from deepening a
single coarse-to-fine chain.

\begin{figure}[t]
\centering
\includegraphics[width=\linewidth]{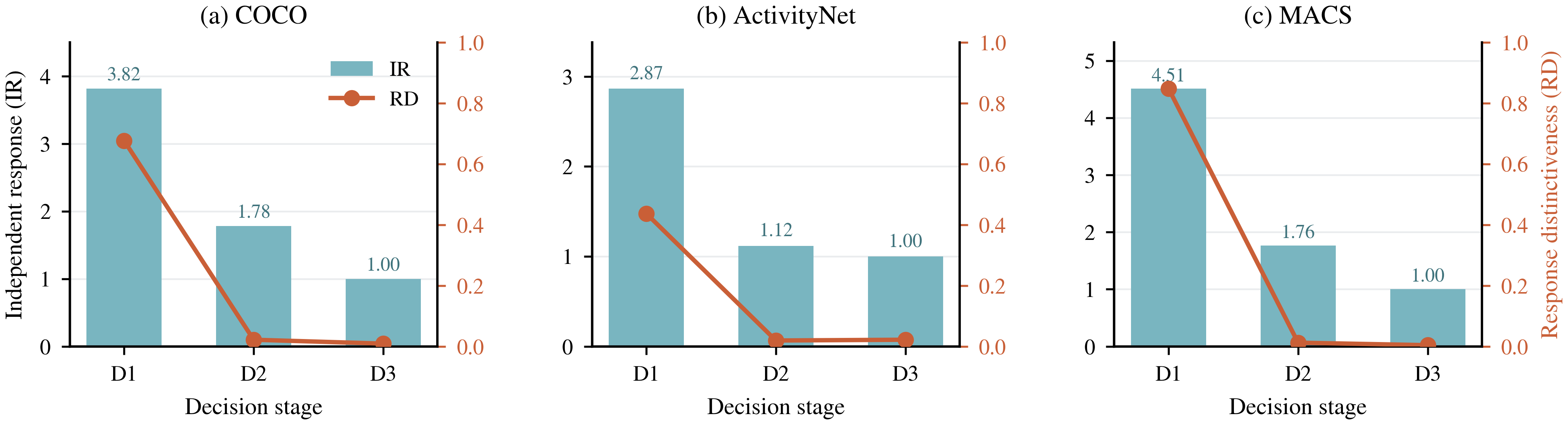}
\caption{
Decision complexity analysis of the hierarchical adaptive policy.
The bars show the average number of independent responses required at each
decision stage, while the curves indicate the remaining response
distinctiveness after each stage. Early decisions provide coarse capacity
separation, allowing later stages to operate on a reduced candidate space.
}
\label{fig:decision_stage}
\end{figure}

\paragraph{Decision Complexity of the Hierarchical Policy.}
\label{app:decision_complexity}
Figure~\ref{fig:decision_stage} analyzes how allocation difficulty is
distributed over the three decision stages. For each stage, the bars report
the average number of independent responses required, and the curves report
the remaining response distinctiveness after the stage. Both quantities
drop sharply after the first decision; on COCO, the required independent
responses fall from $3.8$ to $1.8$, and the residual distinctiveness becomes
substantially weaker. The first coarse decision therefore resolves most of
the capacity-allocation ambiguity, leaving later stages to operate on a
much-reduced candidate space. This concentration of difficulty in the early
stage is what renders the short, three-step hierarchical policy practical:
the expensive discrimination is performed once, at the stage where the
bank-feedback signal is strongest.
\begin{figure}[t]
\centering
\includegraphics[width=\linewidth]{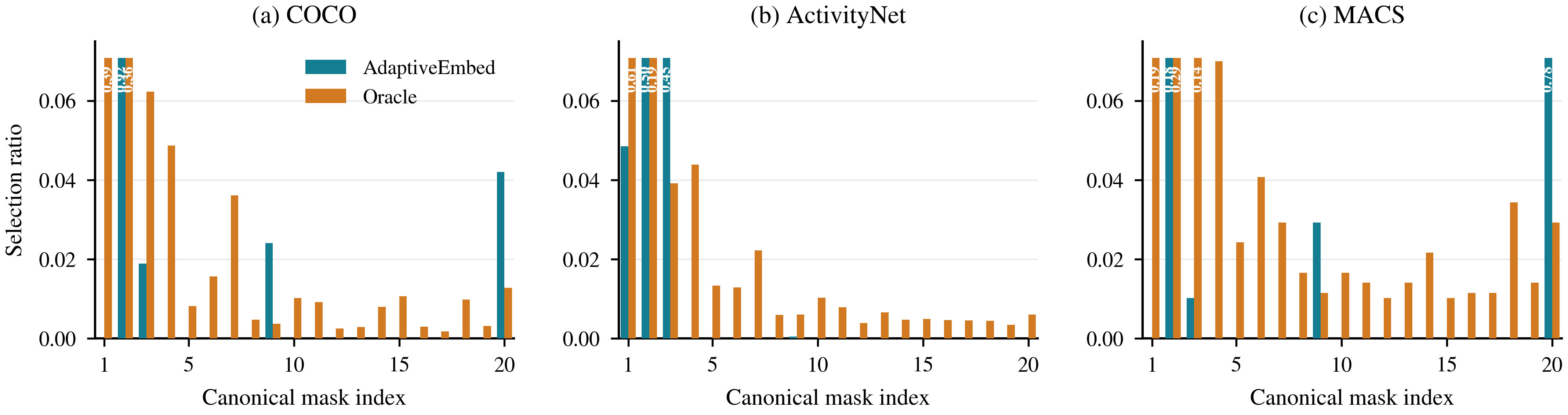}
\caption{
Oracle and AdaptiveEmbed selection distributions on COCO, ActivityNet, and
MACS. Each configuration is represented as a binary mask over the eight
candidate tokens indicating which are activated; the valid group prefixes
yield 20 canonical masks in total. The oracle exhaustively evaluates all 20
masks per sample, whereas AdaptiveEmbed approximates the optimal allocation
through its hierarchical policy.
}
\label{fig:oracle_mask_distribution}
\end{figure}
\paragraph{Learned vs.\ Oracle Allocation Distributions.}
\label{app:oracle_distribution}
Figure~\ref{fig:oracle_mask_distribution} compares the selection
distribution of the learned policy with that of the oracle over the $20$
canonical configurations of $\mathcal{M}$ defined in
Appendix~\ref{app:oracle}, evaluated on COCO, ActivityNet, and MACS. Two
observations stand out. First, the oracle distribution is spread over many
configurations on every dataset, reconfirming at the level of concrete
masks that no single capacity serves all samples. Second, the learned
policy concentrates its mass on a small subset of the oracle's preferred
configurations, primarily the low-capacity masks and the full set, and
rarely selects the intermediate ones. This coarser coverage mirrors the
restricted five-configuration action space of the deployed policy described
in Appendix~\ref{app:decision}, and localizes the policy--oracle gap
observed in Tables~\ref{tab:main_image_text}
and~\ref{tab:main_video_text}. Closing this gap requires finer decision
granularity over the intermediate masks rather than a larger
representation, and finer granularity in turn places higher demands on the
feedback signal: the raw bank responses that suffice for coarse
distinctions weaken rapidly across stages, as shown in
Figure~\ref{fig:decision_stage}, and discriminating among many intermediate
masks would require a richer learned utility signal, for instance a reward
model trained to predict the marginal benefit of each expansion. We regard
such feedback-driven fine-grained allocation as a natural next step for the
SAMVR setting.

\subsection{Qualitative Examples of Sample-Adaptive Allocation}
\label{app:qualitative}
Figure~\ref{fig:qualitative} visualizes \emph{oracle} CAES on COCO samples,
illustrating what optimal sample-wise allocation looks like in practice and
thereby characterizing the SAMVR setting itself, independently of any
learned policy. Three patterns stand out. First, the optimal capacity spans
the entire admissible range from $k{=}1$ to $k{=}8$: no single operating
point serves all samples, which is precisely the premise of the SAMVR
formulation. Second, the optimal allocation is direction-specific, as
several samples require different token sets for I2T and T2I retrieval,
indicating that capacity demand is a property of the retrieval task rather
than of the sample alone. Third, optimal capacity does not reduce to surface
complexity: the most detailed narrative is resolved with a single coarse
token, while a one-sentence caption of a visually ambiguous scene expands to
near-full capacity. What determines capacity is how much representation a
sample needs to be \emph{distinguished within the gallery}, not how much
content it contains. These cases provide sample-level evidence for the
motivation in Figure~\ref{fig:motivation} and the qualitative counterpart to
the complexity--capacity discussion in Appendix~\ref{app:discussion}.

\section{Discussion and Limitations}
\label{app:discussion}

Does sample-adaptive capacity allocation come for free? It does not, and we
discuss the main limitations of our current instantiation, all of which
concern \emph{AdaptiveEmbed} as one realization of SAMVR rather than the
setting itself.

First, a sizable gap remains between the learned policy and the oracle in
Tables~\ref{tab:main_image_text} and~\ref{tab:main_video_text}. The oracle
attains substantially higher mAP at comparable or lower average capacity,
indicating that our hierarchical, few-step policy captures only part of the
achievable sample-wise allocation. Closing this gap requires finer decision
granularity over the intermediate configurations, which in turn places
higher demands on the feedback signal: the raw MaxSim bank responses that
suffice for coarse decisions weaken rapidly across stages, as shown in
Figure~\ref{fig:decision_stage}, and discriminating among many intermediate
configurations would likely require a richer learned utility signal, such
as a reward model trained to predict the marginal benefit of each
expansion.

Second, the efficiency of AdaptiveEmbed is a \emph{matching}-time benefit
that is realized per sample and therefore depends on the data distribution:
the average activated capacity reflects how many samples the policy resolves
with few tokens, and a distribution dominated by hard samples would narrow
the speed advantage. Moreover, because the active token positions are
determined by the query at retrieval time, each gallery item must still
store its full eight-token embedding set so that the query-selected
positions can be read off. Reducing
gallery-side storage under query-dependent activation remains open.

Third, our allocation is driven purely by retrieval utility and is not tied
to any explicit measure of content complexity, since such complexity is
difficult to quantify with a single scalar. As a result, the correspondence
between a sample's semantic richness and its allocated capacity can only be
examined qualitatively; we provide illustrative cases in
Appendix~\ref{app:qualitative} and leave a principled complexity-aware
formulation to future work.

\end{document}